\documentclass{article}

\usepackage{arxiv}

\usepackage[T1]{fontenc}    
\usepackage{hyperref}       
\usepackage{url}            
\usepackage{booktabs}       
\usepackage{amsfonts}       
\usepackage{nicefrac}       
\usepackage{microtype}      
\usepackage{lipsum}		
\usepackage{graphicx}
\usepackage{natbib}
\usepackage{doi}
\usepackage[table]{xcolor}
\usepackage{acronym}
\usepackage{euscript}
\usepackage{xcolor}
\usepackage{multirow}
\usepackage{float}
\usepackage{mathrsfs}
\usepackage[acronym]{glossaries}
\usepackage{threeparttable}
\usepackage{url}
\usepackage{rotating}
\usepackage{tablefootnote}
\usepackage{graphicx} 
\usepackage{lscape}
\usepackage{pdflscape}
\usepackage{subcaption}

\newacronym{CLIP}{CLIP}{Contrastive Language-Image Pre-Training}
\newacronym{CST}{CST}{Context-Aware Soft Target}
\newacronym{SCOLD}{SCOLD}{context-aware Soft target COntrastive Learning for Leaf Disease identification}
\newacronym{DL}{DL}{Deep Learning}
\newacronym{OOD}{OOD}{Out-of-distribution}
\newacronym{SOTA}{SOTA}{State-of-the-art}
\newacronym{VLMs}{VLMs}{Vision-language models}
\newacronym{UME}{UME}{University of Minnesota Extension}
\newacronym{NLM}{NLM}{National Library of Medicine}
\newacronym{ALIGN}{ALIGN}{A Large-scale ImaGe and Noisy-text embedding}

\title{A Vision-Language Foundation Model for Leaf Disease Identification}

\author{
  \href{https://orcid.org/0000-0003-4927-4822}{\includegraphics[scale=0.06]{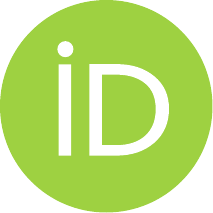}\hspace{1mm}Khang Nguyen Quoc} \\
  School of Electrical Engineering \\
  Korea University, Seoul 02841, South Korea \\
  \texttt{khangnq@korea.ac.kr} \\
  \And
  \href{https://orcid.org/0009-0002-8676-7618}{\includegraphics[scale=0.06]{orcid.pdf}\hspace{1mm}Lan Le Thi Thu} \\
  Department of Software Engineering \\
  FPT University, Can Tho City 900000, Vietnam \\
  \texttt{lanltt@fe.edu.vn} \\
  \And
  \href{https://orcid.org/0000-0002-5661-4250}{\includegraphics[scale=0.06]{orcid.pdf}\hspace{1mm}Luyl-Da Quach} \\
  Department of Software Engineering \\
  FPT University, Can Tho City 900000, Vietnam \\
  \texttt{luyldaquach@gmail.com} \\
}

\renewcommand{\shorttitle}{\textit{arXiv} Template}

\hypersetup{
pdftitle={A template for the arxiv style},
pdfsubject={q-bio.NC, q-bio.QM},
pdfauthor={Khang Nguyen Quoc, Lan Le Thi Thu, Luyl-Da Quach},
pdfkeywords={First keyword, Second keyword, More},
}

\begin{document}
\maketitle

\begin{abstract}
	Leaf disease identification plays a pivotal role in smart agriculture. However, many existing studies still struggle to integrate image and textual modalities to compensate for each other’s limitations. Furthermore, many of these approaches rely on pretraining with constrained datasets such as ImageNet, which lack domain-specific information. We propose SCOLD (Soft-target COntrastive learning for Leaf Disease identification), a context-aware vision-language foundation model tailored to address these challenges for agricultural tasks. SCOLD is developed using a diverse corpus of plant leaf images and corresponding symptom descriptions, comprising over 186,000 image–caption pairs aligned with 97 unique concepts. Through task-agnostic pretraining, SCOLD leverages contextual soft targets to mitigate overconfidence in contrastive learning by smoothing labels, thereby improving model generalization and robustness on fine-grained classification tasks. Experimental results demonstrate that SCOLD outperforms existing vision-language models such as OpenAI-CLIP-L, BioCLIP, and SigLIP2 across several benchmarks, including zero-shot and few-shot classification, image-text retrieval, and image classification, while maintaining a competitive parameter footprint. Ablation studies further highlight SCOLD’s effectiveness in contrast to its counterparts. The proposed approach significantly advances the agricultural vision-language foundation model, offering strong performance with minimal or no supervised fine-tuning. This work lays a solid groundwork for future research on models trained with long-form and simplified contexts, tasks involving class ambiguity, and multi-modal systems for intelligent plant disease diagnostics. The code for this study is available at \url{https://huggingface.co/enalis/scold}.
\end{abstract}


\section{Introduction}
Identifying leaf diseases is critical for precision agriculture and crop protection. Early and accurate detection can optimize yields and minimise large-scale disease outbreaks, contributing to sustainable agricultural practices. This has attracted much research interest in evaluating recently introduced methods, such as soil-based support vector machine algorithms \citep{ref01}, drone-based disease detection \citep{ref02}, hyperspectral imaging-based for quality control of wheat seed \citep{ref58}, and \gls{DL} approaches  \citep{ref03, ref59}. All the evaluation studies have shown that image classification is an important task. However, this problem poses many challenges when utilising text support for image data. Moreover, existing \gls{VLMs} are limited by their reliance on disease-specific datasets, making them ineffective for generalising across multiple crop types. Therefore, this study proposed \gls{VLMs} as leaf-disease classification models for many crops, including the image-text alignment process, to promote the advantages of text and visual features.

The symptoms of plant diseases, especially those expressed in leaves, have aroused wide scholarly interest and have contributed to plant disease classification, which falls into two categories: image and text. The image approach includes many studies on disease classification in tomato \citep{ref04}, rice \citep{ref05}, corn \citep{ref06}, apple \citep{ref07} cultivars, and many other plant types \citep{ref08,ref09}. Large amounts of data from hundreds to tens of thousands of images (such as the PlantVillage dataset) have been collected on multi-crops with accuracy up to $>$96\% but are still constrained by a limited amount of data. Hence, this approach must be tested for more accurate evaluation. In addition, the approach of disease classification based on descriptions/texts has also received a relatively small number of studies. Still, it has also proven feasible when applied to rice diseases \citep{ref10}, swine \citep{ref60} and shrimp \citep{ref11} with relatively high accuracy, up to almost 90\%. Although \gls{DL} models have achieved impressive results in identifying leaf diseases, most studies have focused on disease classification in a single crop, and image-based or text-based approaches are separate. Recent advances in \gls{VLMs} have shown promise in multimodal learning, but their application in disease classification for multiple crops has not been explored.

Although VLMs have developed rapidly in many fields, their application in leaf disease identification remains limited. Some studies have shown success in applying VLMs to solve disease diagnosis problems in wheat \citep{ref12} and potatoes \citep{ref13} by accessing text and images with untrained data types \citep{ref14}, and in cactus cochineal with the transformer-based model \citep{ref61}. However, these studies have limitations owing to the complexity of the natural environment (lighting, viewing angle, background, developmental stages; for example, wheat diseases can show symptoms similar to abiotic stresses, causing classification confusion) and data limitations (images used in laboratory environments, lack of data for rare diseases, data heterogeneity). Current VLMs lack semantic alignment mechanisms, leading to weak text-image correlations and poor feature-learning abilities. Additionally, there is an overfitting and a lack of generalization. Some solutions, such as contrastive learning, have been used to address and solve such problems. However, current models often struggle with limited labelled data, visual similarities between different diseases, and overconfidence in embeddings, which reduces generalisation ability on real-world datasets.

This study proposed and investigated a novel vision-language foundation model named \gls{SCOLD}. \gls{SCOLD} combines contrastive learning and soft targets based on a label-smoothing technique to reduce overconfidence in contrastive learning and improve performance in image-text retrieval tasks. Unlike previous methods that rely solely on image- or text-based classification, \gls{SCOLD} leverages a multimodal learning approach that aligns visual and textual representations through contrastive learning while mitigating overconfidence through label smoothing. This combination enhances robustness, especially in few-shot and zero-shot scenarios. In addition, context-aware adaptation in the training stage benefits the fine-tuning phase by increasing sensitivity to subtle disease features. Finally, large-scale data collection with more than 186,000 images and 97 concepts provides a foundation for further research on leaf disease identification compared to other studies. The research results created a foundation model with \gls{SOTA} performance in multiple benchmarks, specifically downstream vision tasks such as fine-tuning, rather than the pre-trained ImageNet model, with the following main contributions:

\begin{itemize}
     \item The first work applies contrastive learning with \gls{VLMs} to identify leaf diseases.
    \item A newly collected large-scale dataset comprising more than 186,000 images with 97 concepts in an image-text retrieval task provided a comprehensive benchmark.
    \item A novel method was proposed to enhance generalized and clustered feature representations by context-aware label smoothing in contrastive learning.  
    \sloppy\item \gls{SOTA} performance in multiple benchmarks, demonstrating stronger adaptability in few-shot and zero-shot scenarios than conventional methods, such as OpenAI-CLIP-L, BioCLIP, and SigLIP2.
    \item A scalable vision-language model capable of multitasking, providing a foundation for future research and applications in AI-driven smart agriculture.   
\end{itemize}

\section{Related works}
In our discussion of related work, we focus on \gls{VLMs}, contrastive learning, and leaf disease classification.

\subsection{Vision-Language models and contrastive learning}

Recent advancements in \gls{VLMs} have demonstrated strong performance by utilizing contrastive learning techniques to build a unified framework in various tasks. Foundation models such as \gls{CLIP} \citep{ref24} introduced a contrastive learning approach that aligns images and textual descriptions in a shared embedding space, enabling robust zero-shot learning capabilities in common tasks. \gls{ALIGN} \citep{ref28} extended this approach by utilizing large-scale noisy image-text datasets for improved generalization. More recently, FLAVA \citep{ref52} and BLIP \citep{ref53} further advanced multimodal learning by integrating vision and text representations within unified transformer-based architectures.  Despite their success, these models often suffer from high computational costs and complexity due to the huge number of parameters, making them less accessible for domain-specific applications such as agriculture. The reliance on large-scale pretraining datasets and extensive model parameters increases inference time and resource demands, limiting their deployment in real-world agricultural settings. These limitations highlight the need for a more efficient and domain-adaptive approach. The proposed \gls{SCOLD} framework addresses these challenges by enhancing robustness and generalization while maintaining computational efficiency. By leveraging a unified framework as a foundation model for plant disease diagnosis, \gls{SCOLD} reduces complexity and computation time while preserving the effectiveness of contrastive learning in bridging vision and language representations.

\subsection{Leaf disease identification}

Traditional \gls{DL} models have been widely used in leaf disease classification. The PlantVillage \citep{ref29} dataset played a primary role in comparing early CNN-based disease classification models, but this dataset still has limitations such as size and characteristics of images. Moreover, Sajitha \citep{ref54} and Sumaira \citep{ref62} conducted a comprehensive survey on applying deep and machine learning in agriculture, underscoring the importance of large-scale datasets for performance improvements and vision-based methods presents several challenges that need to be addressed. Due to the limited availability of diverse plant disease datasets, many studies have relied on fine-tuning deep learning models pre-trained on ImageNet for simpler cases \citep{ref57}. However, since ImageNet primarily consists of general object categories and biases by uploaders \citep{ref55,ref56}, its domain bias can lead to suboptimal performance when applied to real-world plant disease identification. Developing more comprehensive models with complex architectures is essential for large-scale datasets with diverse and intricate symptoms to improve robustness and generalization. This highlights the urgent need for a sufficiently large-scale dataset encompassing images and textual descriptions to enhance the effectiveness of conventional methods.

\section{Methods}

\subsection{Proposed method}
\begin{figure*}[!ht]
\centering
\includegraphics[width=1.0\textwidth]{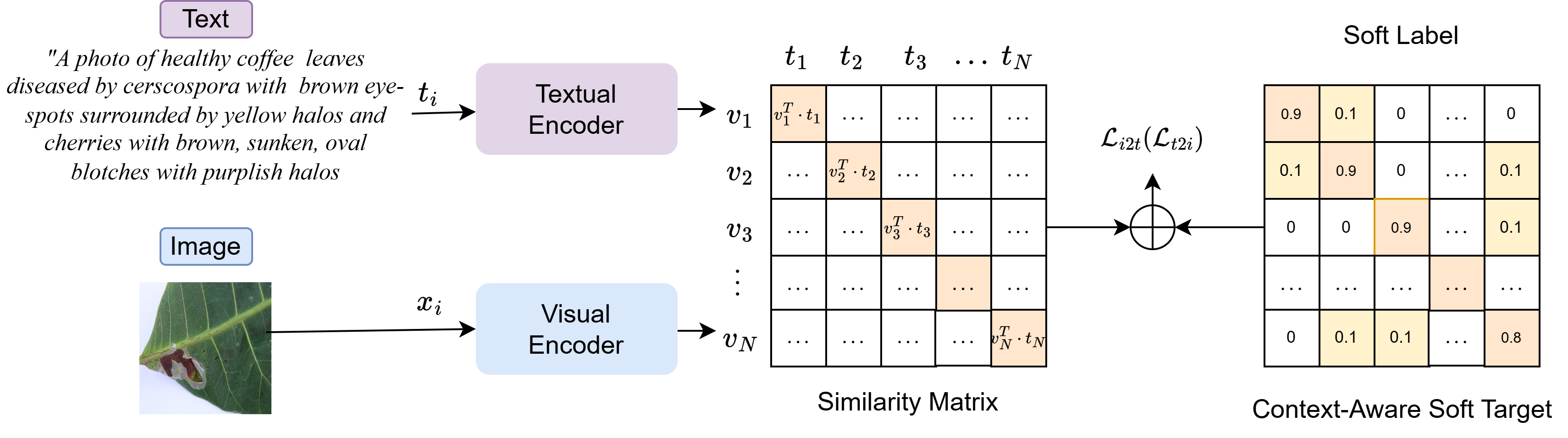}
\caption{The overall framework of \gls{SCOLD}.}
\label{fig1}
\end{figure*}

This study introduced \gls{SCOLD}, a novel vision-language foundation model for context-aware soft-target contrastive language-image for identifying leaf disease. By addressing the challenges of limited image classification and leveraging the rich semantic context provided by textual descriptions, \gls{SCOLD} reformulates the image classification task into an image-text alignment problem by applying contrastive learning with a soft target using context awareness. This shift enables the model to learn more robust and generalisable feature representations, which are particularly effective in downstream tasks like classification and cross-modal retrieval. The overall framework is illustrated in Figure \ref{fig1}.

\subsubsection{Preliminaries}

\sloppy \textbf{Leaf disease identification is an image classification problem} given a set of images and its labels as $D=\left\{ x_i,y_i \right\}_{i=1}^{|D|}$ and $y_i\in \left\{ 0, 1, 2,..., K-1 \right\}$ with $K$ being the total number of classes in $D$, the goal of image classification is to predict the category label of given images accurately using a visual encoder $\mathcal{V}(\bullet )$ and a parametric category such as a softmax classifier $\mathcal{H}(\bullet )$. As the input image $I_i\in \mathbb{R}^{W \times H \times C}$, the $\mathcal{V}(\bullet )$ transforms $I_i$ into an embedding vector $v_i$, with $v_i$ the $\mathcal{H}(\bullet )$ computes the logits distribution $p_i$ overall $K$ categories in $D$. Consider a given image $I_i$, the cross-entropy loss function is used to optimise $p_i$ and $y_i$ is defined as Equation \ref{eq1}. However, \gls{DL} models have many limitations in image classification, such as computing time and feature complexity \citep{ref15}.

\begin{equation}
\label{eq1}
    \mathcal{L}_i=-log\frac{\text{exp}(p_i)}{\sum_{j=1}^{K}\text{exp}(p_i)}
\end{equation}  

\textbf{Image-text alignment} given a set of images and their captions as $D=\left\{ I_i, T_i \right\}_{i=1}^{|D|}$, the goal of image-text alignment is to minimize the distance between matching image-text pairs or positive pairs and increase the distance between non-coincident-pairs, termed negative pairs in the embedding space, through a visual encoder $\mathcal{V}(\bullet )$ and a textual encoder $\mathcal{T}(\bullet )$. The embedding of images and texts $v_i,t_i \in \mathbb{R}^{d}$ in the same d-dimension is transformed by $\mathcal{V}(\bullet )$ and $\mathcal{T}(\bullet )$ after passing through a separate feedforward neural network layer and $l_2$ normalizing. InfoNCE \citep{ref16}, a contrastive loss function is usually applied to adapt the cosine distance of $v_i$ and $t_i$ as in Equation \ref{eq2}, where $sim\left( \bullet, \bullet  \right)$ denotes a similarity function such as dot product and cosine similarity, and the temperature parameter $\tau$ is 0.07.

\begin{equation}
\label{eq2}
    \mathcal{L}_{InfoNCE}=-log\frac{\text{exp}\left( \frac{sim(v_i,t_i)}{\tau} \right)}{\sum_{j=1}^{N}\text{exp}\left( \frac{sim(v_i,t_j)}{\tau} \right)}
\end{equation} 

\begin{figure*}[!ht]
\centering
\includegraphics[width=1.0\textwidth]{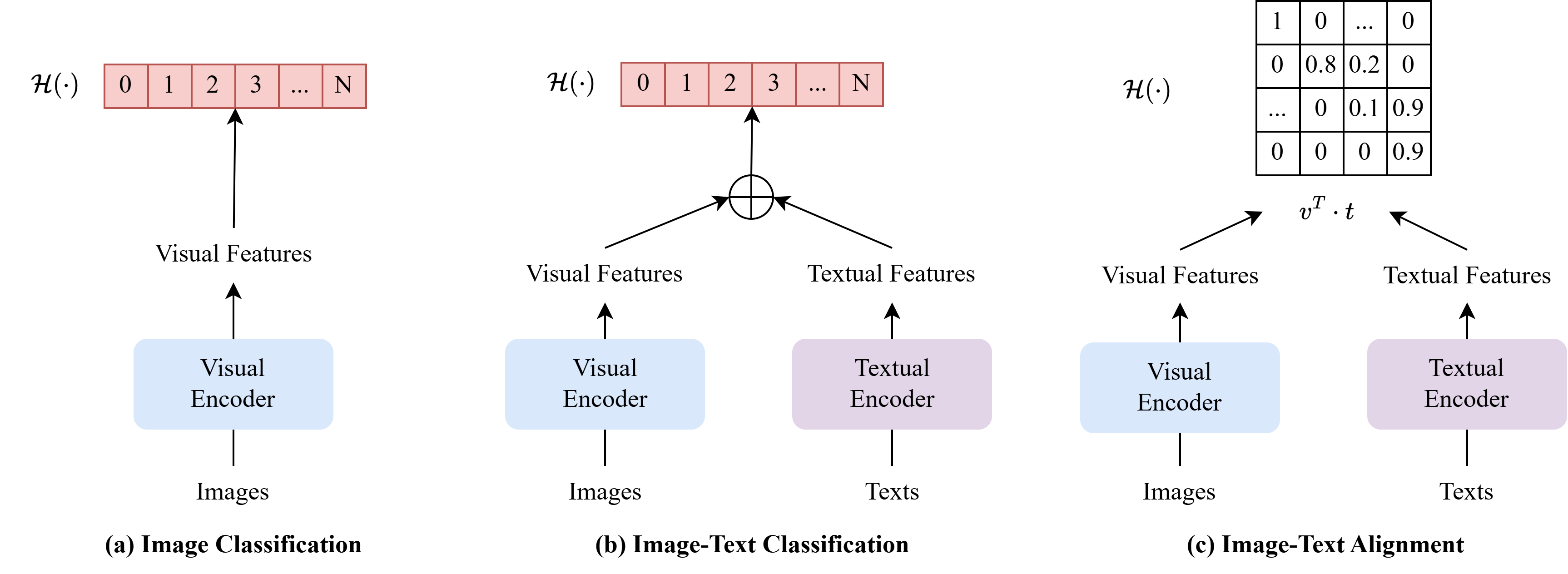}
\caption{The unified model training in \gls{SCOLD} in (c) as Image-Text Alignment with $\mathcal{H}(\bullet )$ as a similarity matrix (white colour) of image-to-text or text-to-image. The limitation problem is that (a) Image classification and (b) Image-Text Classification using $\mathcal{H}(\bullet )$ is constant for each model(red colour).}
\label{fig2}
\end{figure*}

\textbf{Problem Setup} to solve image classification as image-text alignment, we define a triplet-wise data format $S=\left\{ x_i,t_i,y_i|x_i,y_i\in D \right\}_{i=1}^{|S|}$  and $t_i$ is the corresponding textual description. In conventional leaf disease classification, simple category labels or indices $y_i$ are assigned to images. In this task, $t$ text descriptions are $C$ as the set of concept names indexed by $y_i$, and $S$ can be seen as $\left\{ x_i,t_i\equiv C\left[ y_i \right],y_i \right\}$. Figure \ref{fig2} shows an example of their unification. Using the $S$ dataset, image classification is reformulated as an optimisation process, where a pair $(x,t)$ with the same $y$ achieves the highest similarity while being distinct from pairs $(x,t)$ with a different $y$. This study aimed to learn from joint data $S$, believing that the rich semantics in language and structured organisations of the category are beneficial for feature representation in downstream tasks, specifically in image classification.

\subsubsection{Context-aware Soft-target Contrastive Language-Image Pre-Training for Leaf Disease Identification}

Many studies have used multimodal or vision-language models to improve image classification, including models that have taken advantage of the combination of text and image features, such as iCar \citep{ref17}, UniCL \citep{ref18}, and iCLIP \citep{ref19}. Hence, using many different approaches, we present a vision-language model for building a unified classifier and a better extractor for downstream tasks in smart agriculture problems with four proposed approaches are textual encoder as a novel network for classification, contrastive language-image pre-training, enriching class concepts and context-aware soft target.

\textbf{a) Textual encoder as a novel network for leaf disease classification}

Many studies have shown that text features greatly enhance vision models \citep{ref20}. However, the application of vision-language models in agriculture is still limited, especially because leaf disease classification is based only on visual features and focuses on a single crop, e.g., tomatoes \citep{ref21}, maize \citep{ref22}, or cucumbers \citep{ref23}. Integrating textual encoders provides rich concepts and the open-vocabulary ability for multi-task learning, such as image-text retrieval and image classification, using the $S$ format and $\mathcal{T}(\bullet )$ encoder, $t$ text description by $C[y_i]$ concept with its corresponding class name, for example, \textit{Apple Scab} for the $1^{st}$ category in the PlantVillage dataset \citep{ref29}. To this end, with the information learned from text descriptions, the text encoder enables the open-vocabulary ability in agriculture and builds a knowledge foundation from diverse crop disease information.

\textbf{b) Contrastive Language-Image Pre-Training}
We present a vision-language model based on the concept of \gls{CLIP}\citep{ref24} to align leaves' textual and visual features in representation spaces. As in the \gls{CLIP} setup, given a batch with $N$ image-text pairs $\left\{ x_i,t_i \right\}_{i=1}^N$, with dual-stream encoders for visual $\mathcal{V}(\bullet )$ and the textual encoder $\mathcal{T}(\bullet )$. To compute the semantic representation of each pair, the embedding $v_i$ and $t_i$ is transformed by $\mathcal{V}(\bullet )$ and $\mathcal{T}(\bullet )$ and computed by a different feedforward neural network with an output size of $d=512$. This embedding generates $l_2$ normalised embedding of each image-text pair. As mentioned in Equation \ref{eq2}, the corresponding one-hot label vectors are used as the targets to calculate losses, including image-to-text loss and text-to-image loss as defined in Equations \ref{eq3} and \ref{eq4}, with the label of $i^{th}$ pair being $y_{i}=\left\{ y_{ij} \right\}_{i=1}^N$, with paired one. Have $y_{ij}$ equal one and another negative pair be 0. Therefore, the final loss in the \gls{CLIP} model is $\mathcal{L}_{CLIP}=\frac{\mathcal{L}_{i2t}+\mathcal{L}_{t2i}}{2}$.

\begin{equation}
\label{eq3}
    \mathcal{L}_{i2t}=\frac{1}{N}\sum_{i=1}^{N}R\left( y_i,\mathcal{L}_{InfoNCE}\left( \mathcal{V}{(x_i)},\mathcal{T}(t_i) \right) \right)
\end{equation} 

\begin{equation}
\label{eq4}
    \mathcal{L}_{t2i}=\frac{1}{N}\sum_{i=1}^{N}R\left( y_i,\mathcal{L}_{InfoNCE}\left( \mathcal{T}(t_i),\mathcal{V}{(x_i)} \right) \right)
\end{equation} 
where $R\left( \bullet, \bullet  \right)$ denotes the cross-entropy operation.

\textbf{c) Enriched class concepts with descriptions of diseases and symptoms} 

\begin{figure*}[!ht]
\centering
\includegraphics[width=0.8\textwidth]{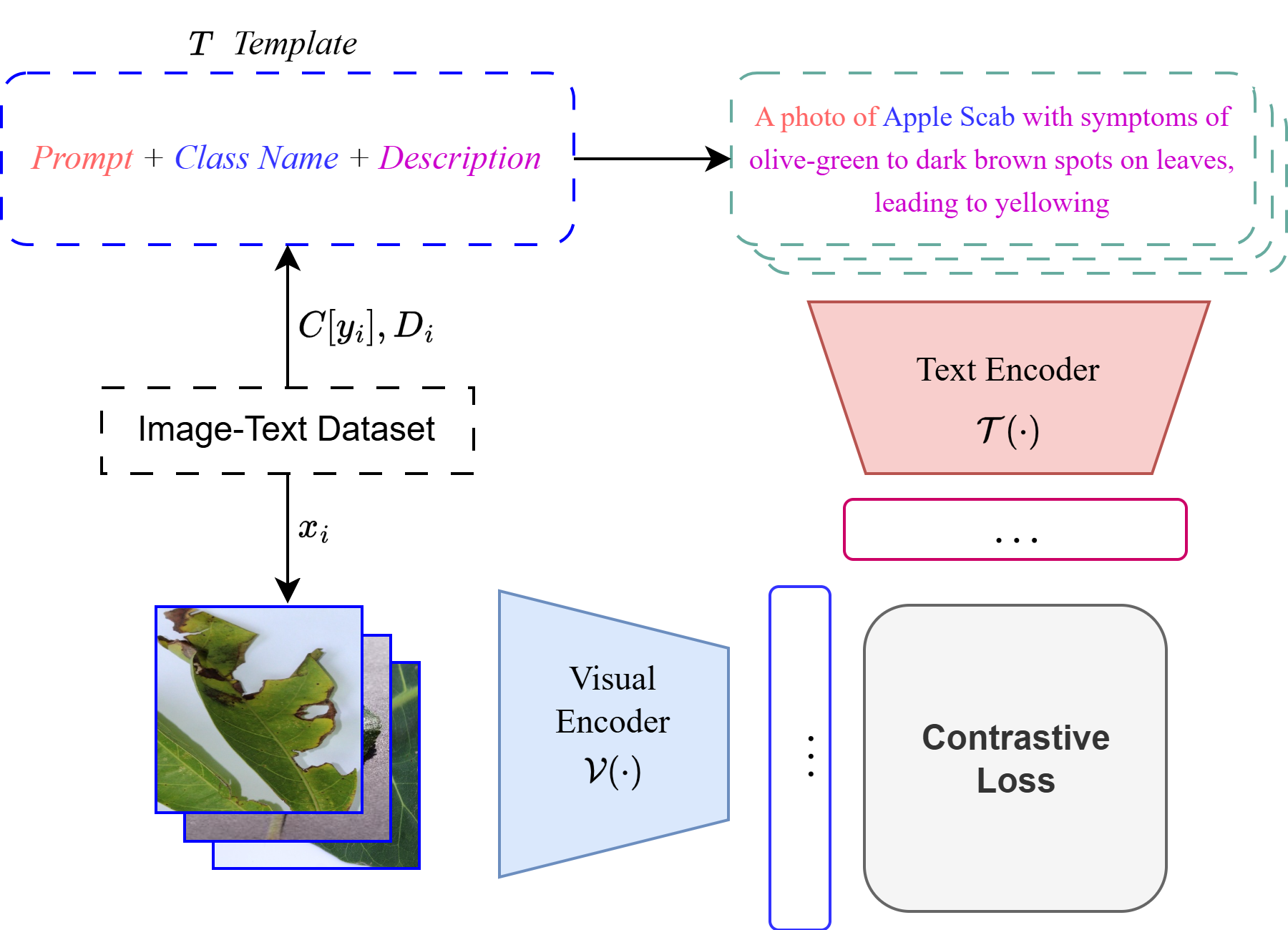}
\caption{A detailed illustration of image-text alignment with a prompt template for enriching class information with a common plant-leaf disease dataset such as PlantVillage \citep{ref29}.}
\label{fig3}
\end{figure*}

This method uses text descriptions instead of class indices, which are needed for information diversity between images and text. However, only using $C[y_i]$ concepts based on the class name (a noun or pronoun) instead of image caption (a comprehensive sentence) concepts will introduce certain limitations. Therefore, we integrated a detailed description of $D_i$ for each disease. These descriptions summarise the disease symptom information from the \gls{UME}\footnote{https://extension.umn.edu/}  and the \gls{NLM}\footnote{https://pubmed.ncbi.nlm.NLM.gov/}. To improve the fluency, we added a prompt template. As illustrated in Figure \ref{fig3}, the class name is also concept $C$ for each category, which is formed via the following the $T$ template as follows: \textit{prompt +[crop name]+leaves diseased by+[disease name]+with symptoms of+[description]}, but for the healthy class, we use another; \textit{prompt +[crop name]+healthy leaves with leaves appearing normal and healthy}, the final text description $t_i$ for Equations \ref{eq3} and \ref{eq4}. 

A detailed description provides a more insightful understanding of each disease and shares feature information among the crop categories and disease symptoms. Moreover, this can reduce misconception errors in limited contexts, such as using only class names. For example, the PlantVillage dataset contains 38 classes \citep{ref29}. However, \textit{Blight} is represented by six different classes in many types of plants. Therefore, using a long context as a detailed description with its class name has granularity similar to that of captions, thus making image–text alignment easier.

\textbf{d)Context-Aware Soft Target}
As mentioned, the features learned from each leaf disease dataset often overlap or are similar. This limits the \gls{CLIP} model because it prevents features belonging to negative pairs in the same batch from being processed. We proposed a Softened Target to overcome the limitations of one-hot labels in \gls{CLIP}. For instance, label smoothing \citep{ref26}, a common technique in classification tasks, involves assigning small values to all negative samples. Moreover, SoftCLIP applies a Softened Target to employ a fine-grained intramodel to solve the nonstrict mutual exclusion problem between any two pairs \citep{ref27}. 

\begin{figure*}[!ht]
\centering
\includegraphics[width=0.8\textwidth]{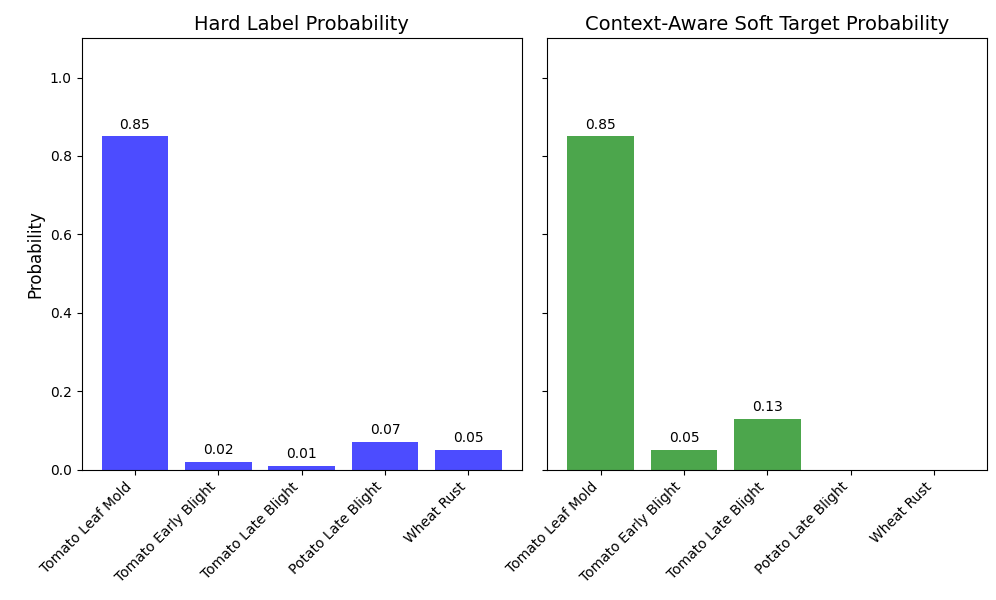}
\caption{Disentangling the negatives in similar distributions. With an image query Tomato Leaf Mould, the probability distribution of the hard label can tend to appear in different crop classes, such as Potato or Wheat, due to distance-only optimization. Meanwhile, with \gls{CST}, this probability distribution will be distributed into the most similar classes to the query.}
\label{fig4}
\end{figure*}

In this study, we propose a \gls{CST} as a crop and disease-type label smoothing strategy to perform downstream tasks better and avoid this effect. With \gls{CST}, the proposed \gls{SCOLD} model will be more efficient and reliable in downstream tasks, simultaneously mitigating the problem, and more serious when encountering faulty positives, paired images and texts from various resources. Figure \ref{fig4} illustrates the results of the conventional \gls{CLIP} and \gls{CST} techniques after adaptation.

\sloppy To implement soft-target, we use label smoothing for all pairs in the batch based on \textit{crop name} and \textit{disease name} to form a soft label matrix $P\in \mathbb{R}^{|B|\times |B|}$, where $c$ represents the type of plant, $d$ represents the disease type from the $C$ concept, and $\rho_k$ is the label value of the $k^{th}$ in $P$ label matrix (Equation \ref{eq5}).

\begin{equation}
\label{eq5}
    \rho_k=\left\{ \begin{array}{cl}
1-\alpha-\beta, & if\ c_k=c_i\ and\ d_k=d_i\\
\frac{\alpha}{N_{crop}}, & if\ c_k=c_i\ and\ d_k\neq d_i\\
\frac{\beta}{N_{disease}}, & if\ c_k\neq c_i\ and\ d_k= d_i\\
0, & otherwise\\
\end{array} \right.
\end{equation} 



\begin{figure*}[!ht]
\centering
\includegraphics[width=1.0\textwidth]{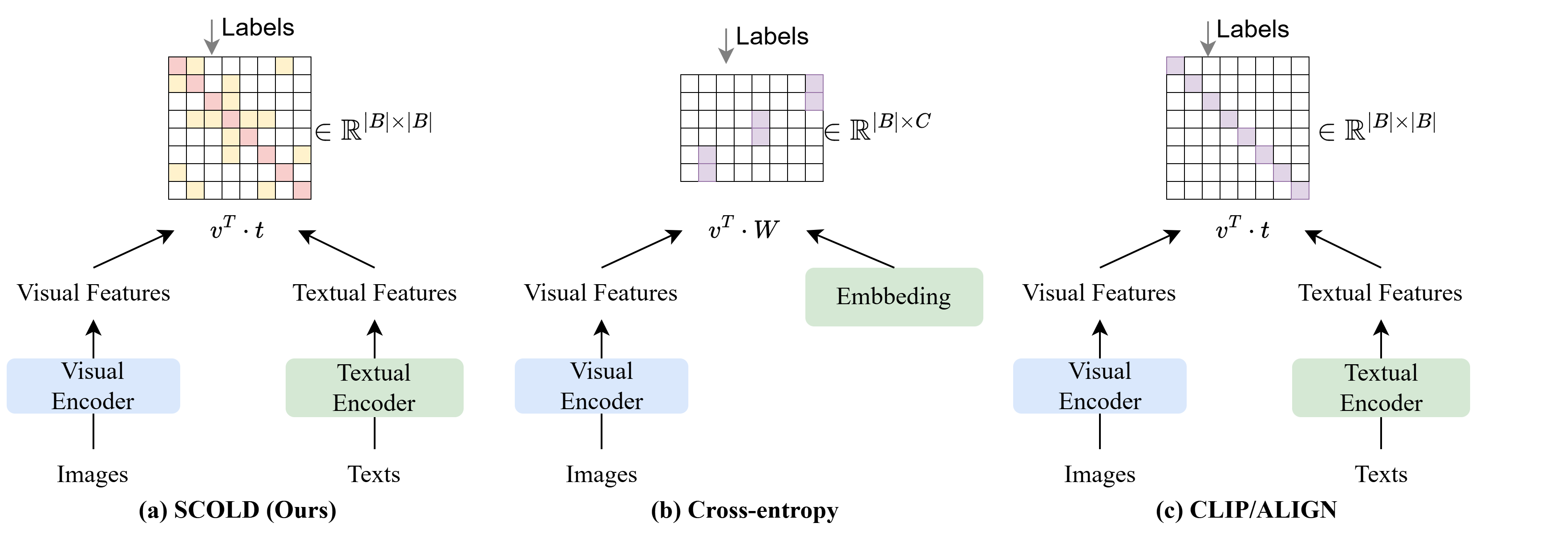}
\caption{Illustrative comparisons across different learning paradigms. With a batch of size $|B|$, all visual features $v$ and textual features $t$ are in the same dimension, and $C$ is the number of classes. Given a similarity matrix for each method, the labels are defined as the positive pairs whose elements are purple and whose negatives are white. (c) \gls{CLIP} \citep{ref24} or \gls{ALIGN} \citep{ref28} has a one-to-one assumptions for each pair. Meanwhile, (a) \gls{SCOLD} has many-to-many assumptions based on the context; positive pairs are considered based on orange, related pairs are negative yellow, and the remaining non-related pairs are white.}
\label{fig5}
\end{figure*}

With $N_{crop}$ and $N_{disease}$, several other samples with the same crop and disease are compared similarly to the positive pairs in the batch. Following this, the pair with 0 is the negative pair, the other is a related pair for the crop and disease, and the $\alpha$ and $\beta$ control its strength. We use $\alpha=0.1$ and $\beta=0.05$ for the training setup. Based on this, the infoNCE loss in Equation \ref{eq2} is adjusted as in Equation \ref{eq6} to optimise the \gls{SCOLD} model. Figure \ref{fig5} shows the label matrices of certain learning paradigms and the proposed model.

\begin{equation}
\label{eq6}
\mathcal{L}_{InfoNCE}=-\frac{1}{N}\sum_{i=1}^{N} \rho_k. log\frac{\text{exp}\left( \frac{sim(v_i,t_i)}{\tau} \right)}{\sum_{j=1}^{N}\text{exp}\left( \frac{sim(v_i,t_j)}{\tau} \right)}
\end{equation} 

The problem of computation time and resources when training vision-language models is complex; therefore, the study proposes a batch sampler for the training process. This batch sampler is responsible for each image-text pair belonging to different classes, which helps the model to work well for small batch sizes, while large batch sizes are often preferred for training \gls{CLIP} models.

\section{Experiments Results}
\subsection{Dataset}
To our knowledge, the PlantVillage dataset is the largest currently available in the leaf disease classification domain, with over 54,000 images of 14 plant species with 17 fungal diseases, four bacterial diseases, two mould (oomycete) diseases, two viral diseases, and one mite disease \citep{ref29}. However, this dataset is limited in both features and quantities. Therefore, we collected additional data from various studies to create a large dataset for the foundation model for leaf diseases. 

\begin{figure*}[!ht]
 \centering
 \includegraphics[width=1.0\textwidth]{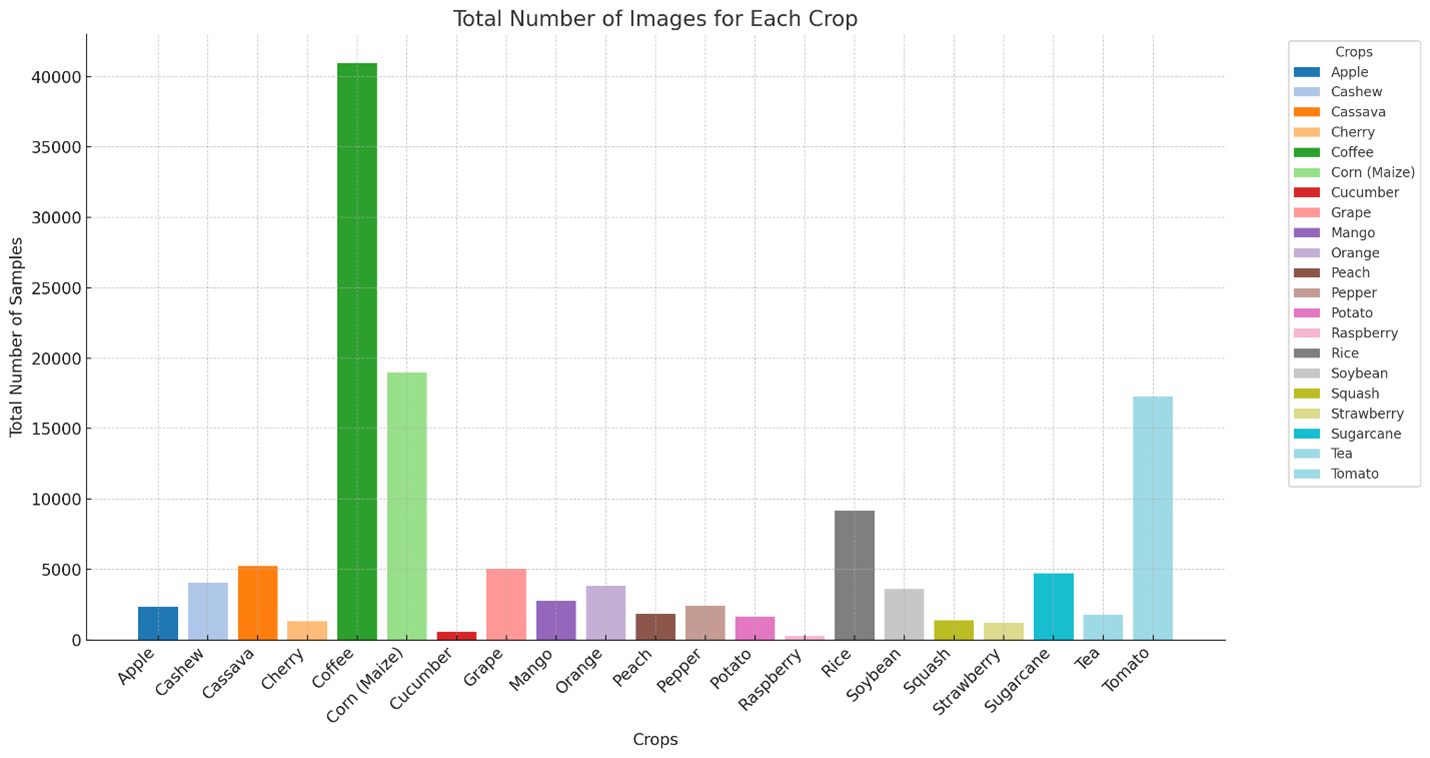}
 \caption{Distribution of images for each crop with leaf diseases.}
 \label{fig6}
\end{figure*}

After careful collection and screening, we assembled a new dataset named LeafNet comprising $>$186,000 images of 22 crop species (Figure \ref{fig6}) with 43 fungal diseases, eight bacterial diseases, two moulds diseases, six viral diseases, and three diseases caused by a mite, and divided into 97 classes. The above data were collected and processed to ensure that the leaf characteristics in the dataset did not vary much between classes and only contained images taken at a certain distance for clarity. The sources of the collected information and data distributions are shown in Table \ref{tab1}, and LeafNet was divided into training, validation, and testing sets at 7:2:1.

\begin{table}
\centering
\begin{threeparttable}

\caption{Dataset resources.}
\label{tab1}
\begin{tabular}{|p{7cm}|r|r|}
\hline
\textbf{Dataset} & \textbf{Total Class} & \textbf{Total images} \\ \hline
PlantVillage\citep{ref29} & 38 & 54,303 \\ \hline
Grape Leaf dataset\citep{ref30} & 5 & 3,000 \\ \hline
Sugarcane Leaf Dataset\citep{ref31} & 11 & 6,748 \\ \hline
MangoLeafBD Dataset\citep{ref32} & 8 & 4,000 \\ \hline
JMuBEN\citep{ref33} & 5 & 58,555 \\ \hline
Black Pepper Leaf Blight and Yellow Mottle Virus\tablefootnote{Dataset available at \url{https://surl.li/ktxrzi}} & 3 & 273 \\ \hline
Tea Crop Disease\citep{ref34} & 4 & 2,520 \\ \hline
\sloppy Crop Pest and Disease Detection\citep{ref35} & 22 & 24,481 \\ \hline
Maize dataset\citep{ref36} & 3 & 18,148 \\ \hline
Cucumber dataset\citep{ref37} & 5 & 4,000 \\ \hline
Rice Leaf Disease\citep{ref38} & 8 & 13,186 \\ \hline
PlantDoc\citep{ref39} & 17 & 2,598 \\ \hline
\end{tabular}

\end{threeparttable}
\end{table}

\subsection{Pre-training Model}
\textbf{The \gls{SCOLD} architecture:} supports the low-setting model architecture with a visual encoder compatible with Swin-T \citep{ref40}, whereas the language encoder leverages RoBERTa \citep{ref41}, following the \gls{CLIP} framework \citep{ref24}. The input size of the visual encoder was $224 \times  224$, and the maximum context length of the textual encoder was 77. Finally, \gls{SCOLD} was pre-trained on the LeafNet dataset.

\textbf{Training Setup:} We trained our \gls{SCOLD} model using an AdamW \citep{ref42} optimizer and a linear-learning rate scheduler with a linear warm-up. This means that the learning rate linearly increases from 0 to 10\% of the peak value of the total step and then decreases with a cosine annealing strategy. For the training set, the weight decay was 0.2. The model was trained from scratch for 20 epochs. An NVIDIA GeForce RTX 3060 Ti 12 GB was used for the experimental hardware setup.
\subsection{Comparison Setup}

We conducted comprehensive experiments to assess the task-specific capabilities of SCOLD. Specifically, evaluations were performed across ten distinct subsets encompassing diverse vision-language tasks, including image classification and image-text retrieval.

\textbf{Evaluation Details:} Our experiments were conducted on ten \gls{OOD} datasets: the Wheat Leaf dataset \citep{ref43}, the Chili leaf dataset, the Onion leaf dataset \citep{ref44}, the Durian Leaf Disease dataset\footnote{https://universe.roboflow.com/new-workspace-7ly0p/durian-diseases}, SouropDB \citep{ref45}, Radish \citep{ref47}, Black gram \citep{ref48}, Eggplant \citep{ref49}, Guava \citep{ref50}, and Spinach \citep{ref51} (Table \ref{tab21}). Our model’s performance was benchmarked against OpenAI-CLIP-L \citep{ref24} with fine-tuning on LeafNet, SigLIP2 \citep{ref65}, ResNet50V2 \citep{ref63} and \gls{SOTA} BioCLIP model \citep{ref64} in plant sciences (Table \ref{tab:model_comparison}).


\begin{table}[htbp]
\centering
\caption{Comparison of vision-language models used for few-shot leaf disease classification. SCOLD leverages domain-specific pretraining on the LeafNet dataset.}
\label{tab:model_comparison}

\begin{tabular}{|p{2.2cm}|p{2cm}|p{2.7cm}|p{2.85cm}|p{1.8cm}|}
\hline
\textbf{Model} & \textbf{Vision Encoder} & \textbf{Text Encoder} & \textbf{Pretraining Data} & \textbf{\#Params (M)} \\
\hline
OpenAI-CLIP-L     & ViT-L/14         &  Masked Self-attention Transformer               & LAION-2B                       & 428 \\
BioCLIP        & ViT-B/16         & Causal Autoregressive transformer           & TREEOFLIFE-10M                 & 149 \\
SigLIP2        & EfficientNet-B7  & Sigmoid Text Encoder & JFT-4B + CC3M                  & 375 \\
\rowcolor{cyan!10}
SCOLD (Ours)   & Swin-T           & RoBERTa-base         & LeafNet                        & 171 \\
\hline
\end{tabular}
\end{table}

\textbf{Evaluation Tasks:} We evaluated the image-text alignment and feature extraction capabilities of \gls{SCOLD} through zero-shot classification, few-shot linear probing and fine-tuning, image classification and image-text retrieval.

For few-shot linear probing and fine-tuning, we added a linear layer to extracted feature representations from four datasets: Black gram, Chili, Onion and Radish datasets. For linear probing, training sizes of 1, 4, 16, 32, 64, and 128 shots and each size was randomly sampled 10 times for 10 runs. Beside, fine-tuning training sizes are 1, 4 and 16 shots. The variability and robustness of the model’s performance is illustrated by box plots.

\begin{table*}
\begin{center}
\caption{\gls{OOD} datasets description.}
\label{tab21}
\begin{tabular}{|l|r|r|}
\hline
\multicolumn{1}{|c|}{\textbf{Dataset}} & \textbf{Total Class} & \multicolumn{1}{c|}{\textbf{Total images}} \\ \hline
Black gram                             & 5                    & 4,038                                      \\ \hline
Chili                                  & 5                    & 10,794                                     \\ \hline
Durian                                 & 4                    & 413                                        \\ \hline
Eggplant                               & 6                    & 4089                                       \\ \hline
Guava                                  & 5                    & 2,065                                      \\ \hline
Onion                                  & 4                    & 816                                        \\ \hline
Radish                                 & 5                    & 2,081                                      \\ \hline
Soursop                                & 6                    & 3,838                                      \\ \hline
Spinach                                & 5                    & 3,006                                      \\ \hline
Wheat                                  & 3                    & 407                                        \\ \hline
\end{tabular}

\end{center}
\end{table*}

\subsection{Zero-shot Classification}
SCOLD demonstrated superior zero-shot classification performance across the majority of leaf disease \gls{OOD} datasets. As shown in Table \ref{tab3}, SCOLD notably outperformed the current state-of-the-art models, including CLIP, SigLIP2, and BioCLIP, with the highest average accuracy of 34.80\%. SCOLD achieved the best results on 6 out of 10 datasets, including Wheat (42.26\%), Chili (17.22\%), Onion (62.33\%), Eggplant (36.64\%), Guava (59.24\%), and Spinach (48.39\%), and ranked second on the Radish (25.73\%) dataset. Compared to the closest competitor, SigLIP2, SCOLD improved average accuracy by a margin of 0.88\%. These results highlight SCOLD's enhanced ability to align image and text features effectively, particularly in the domain of plant disease recognition.
\begin{landscape}
\begin{table}
    \small
    \centering
    \caption{Zero-shot classification comparison of various models on different leaf disease image classification \gls{OOD} datasets with accuracy (\%). The best performance is highlighted in \textbf{bold}, while the second-best is \textit{italic}. Abbreviations: W $=$ Wheat, C $=$ Chili, D $=$ Durian, O $=$ Onion, S $=$ Soursop, BG $=$ Black gram, EG $=$ Eggplant, GV $=$ Guava, RD $=$ Radish, SP $=$ Spinach and Avg. $=$ Average.}
    \label{tab3}
    \begin{tabular}{|l|c|c|c|c|c|c|c|c|c|c|c|}
    \hline
    \textbf{Model} 
    & \textbf{W} 
    & \textbf{C} 
    & \textbf{D} 
    & \textbf{O} 
    & \textbf{S} 
    & \textbf{BG} 
    & \textbf{EP} 
    & \textbf{GV} 
    & \textbf{RD} 
    & \textbf{SP} 
    & \textbf{Avg.} \\
    \hline
    \gls{CLIP}
    & 27.02
    & 11.87
    & \textit{60.01}
    & 16.91
    & 16.57
    & 31.28
    & \textit{33.41}
    & 51.19
    & 24.66
    & 46.04
    & 31.70 \\
    \hline
    SigLIP2
    & \textit{31.07}
    & 13.52
    & \textbf{62.33}
    & \textit{16.91}
    & \textit{18.21}
    & \textit{33.18}
    & 30.55
    & \textit{53.86}
    & 25.71
    & \textit{48.06}
    & \textit{33.92} \\
    \hline
    BioCLIP 
    & 25.35
    & \textit{14.33}
    & 41.33
    & 11.23
    & \textbf{20.85}
    & \textbf{38.12}
    & 25.15
    & 49.53
    & \textbf{28.64}
    & 39.06
    & 29.66 \\
    \hline
    \rowcolor{cyan!10} 
    \gls{SCOLD}
    & \textbf{42.26}
    & \textbf{17.22}
    & 46.97
    & \textbf{20.39}
    & 17.11
    & 32.01
    & \textbf{36.64}
    & \textbf{59.24}
    & \textit{25.73}
    & \textbf{48.39}
    & \textbf{34.80} \\
    \hline
    \end{tabular}
\end{table}
\end{landscape}

\subsection{Few-shot Classification}
In the few-shot fine-tuning setting, \gls{SCOLD} consistently outperformed baseline models, showcasing strong generalization capabilities across diverse plant disease classification tasks (Table \ref{tab4}). In the 1-shot scenario, SCOLD achieved the highest accuracy on five datasets-Eggplant (46.09\%), Guava (66.62\%), Soursop (82.64\%), Spinach (48.32\%), and Wheat (67.32\%)-and secured second-best results on Durian (78.93\%) and Onion (43.38\%). At 4-shot, \gls{SCOLD} led on four datasets, notably Black gram (90.09\%) and Spinach (62.45\%), with competitive gains across the rest. In the 16-shot setting, it achieved top performance on six datasets-Black gram (97.22\%), Durian (97.09\%), Eggplant (54.55\%), Guava (84.22\%), Radish (95.82\%), and Spinach (60.81\%)-demonstrating its robustness and effectiveness even with limited training data.

\begin{landscape}
\begin{table*}[htbp]
\centering
\caption{Few-shot classification comparison of various fine-tuning models on different leaf disease image classification datasets with accuracy (\%). The best performance is highlighted in bold, while the second-best is underlined. The models used include ResNet50V2 \citep{ref63}, CLIP\citep{ref24}, BioCLIP \citep{ref64}, SigLIP2 \citep{ref65} and SCOLD (Our proposed). Abbreviations: W $=$ Wheat, C $=$ Chili, D $=$ Durian, O $=$ Onion, S $=$ Soursop, BG $=$ Black gram, EG $=$ Eggplant, GV $=$ Guava, RD $=$ Radish and SP $=$ Spinach.}
\label{tab4}
\renewcommand{\arraystretch}{1.15}
\begin{tabular}{|c|l|c|c|c|c|c|c|c|c|c|c|}
\hline
\textbf{Shot} & \textbf{Model} & \textbf{BG} & \textbf{C} & \textbf{D} & \textbf{O} & \textbf{EP} & \textbf{GV} & \textbf{RD} & \textbf{S} & \textbf{SP} & \textbf{W} \\
\hline
\multirow{5}{*}{1-shot}
& ResNet50V2     & 53.33 & 37.56 & 39.12 & 29.84 & 22.26 & 48.59 & 47.17 & 61.64 & 37.99 & \textit{66.58} \\
& CLIP           & 57.90 & 48.92 & \textbf{79.22} & \textbf{44.56} & 40.93 & \textit{66.08} & \textbf{69.17} & 80.36 & \textit{40.15} & 65.72 \\
& BioCLIP        & \textbf{73.03} & \textit{50.58} & 51.16 & 27.05 & \textit{43.82} & 48.11 & 53.16 & 71.13 & 22.03 & 54.54 \\
& SigLIP2        & \textit{71.84} & \textbf{52.03} & 58.13 & 35.29 & 38.49 & 44.81 & \textit{62.67} & \textit{80.92} & 22.03 & 59.10 \\
 \rowcolor{cyan!10} & SCOLD & 59.51 & 49.91 & \textit{78.93} & \textit{43.38} & \textbf{46.09} & \textbf{66.62} & 55.22 & \textbf{82.64} & \textbf{48.32} & \textbf{67.32} \\
\hline
\multirow{5}{*}{4-shot}
& ResNet50V2     & 51.39 & 64.72 & 58.69 & 20.78 & 28.51 & 32.60 & 61.52 & 88.10 & 52.48 & 69.87 \\
& CLIP           & \textit{87.21} & \textit{67.85} & \textbf{90.01} & 54.21 & \textbf{54.27} & \textbf{78.27} & 74.15 & 91.41 & \textit{57.10} & \textit{82.11} \\
& BioCLIP        & 79.16 & 56.66 & 74.41 & 29.41 & 41.40 & 46.69 & \textit{81.69} & \textit{92.26} & 26.97 & 81.81 \\
& SigLIP2        & 77.94 & 62.55 & 83.72 & \textbf{58.82} & 38.49 & 51.41 & \textbf{84.51} & \textbf{96.13} & 25.65 & 75.00 \\
\rowcolor{cyan!10}& SCOLD & \textbf{90.09} & \textbf{69.91} & \textit{87.89} & \textit{55.88} & \textit{47.87} & \textit{77.63} & 70.36 & 91.56 & \textbf{62.45} & \textbf{83.04} \\
\hline
\multirow{5}{*}{16-shot}
& ResNet50V2     & 93.81 & \textbf{79.47} & 94.27 & 47.14 & \textit{51.49} & \textit{81.95} & 94.19 & \textbf{99.60} & \textit{59.71} & 79.94 \\
& CLIP           & \textit{95.98} & 69.11 & \textit{96.99} & 61.59 & 48.46 & 81.64 & \textit{95.00} & 96.98 & 57.79 & 85.48 \\
& BioCLIP        & 89.46 & 73.25 & 86.04 & 60.00 & 45.03 & 64.15 & 92.25 & 97.16 & 59.21 & \textit{88.63} \\
& SigLIP2        & 90.03 & \textit{75.24} & 90.06 & \textbf{64.71} & 44.30 & 75.00 & 92.95 & 97.68 & 54.60 & \textbf{90.10} \\
\rowcolor{cyan!10} & SCOLD & \textbf{97.22} & 69.98 & \textbf{97.09} & \textit{62.58} & \textbf{54.55} & \textbf{84.22} & \textbf{95.82} & \textit{98.61} & \textbf{60.81} & 86.73 \\
\hline
\end{tabular}
\end{table*}
\end{landscape}

\begin{landscape}
    
\begin{figure}[htbp]
    \centering

    \begin{subfigure}[b]{0.65\textwidth}
        \centering
        \includegraphics[width=\linewidth]{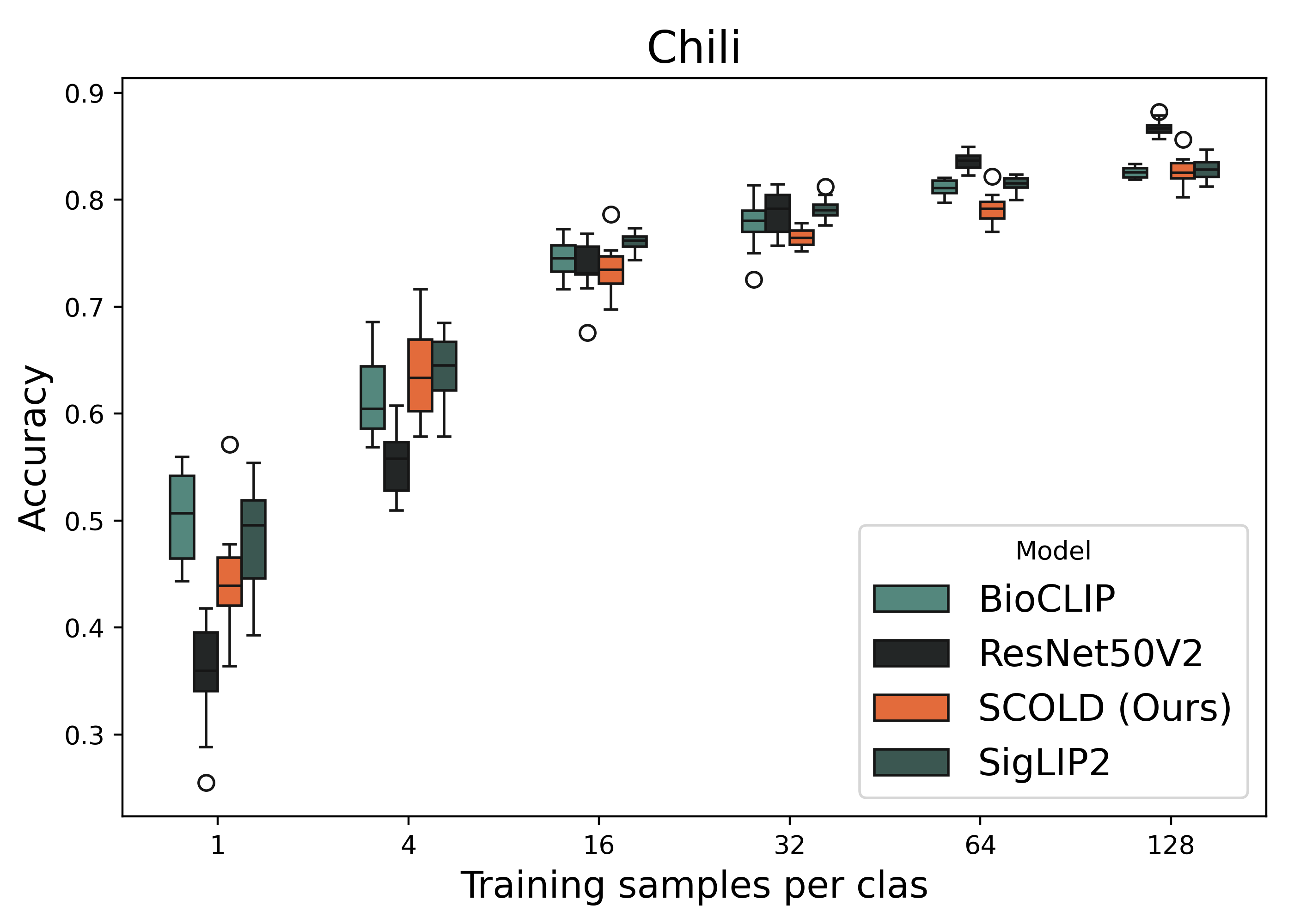}
    \end{subfigure}
    \hfill
    \begin{subfigure}[b]{0.65\textwidth}
        \centering
        \includegraphics[width=\linewidth]{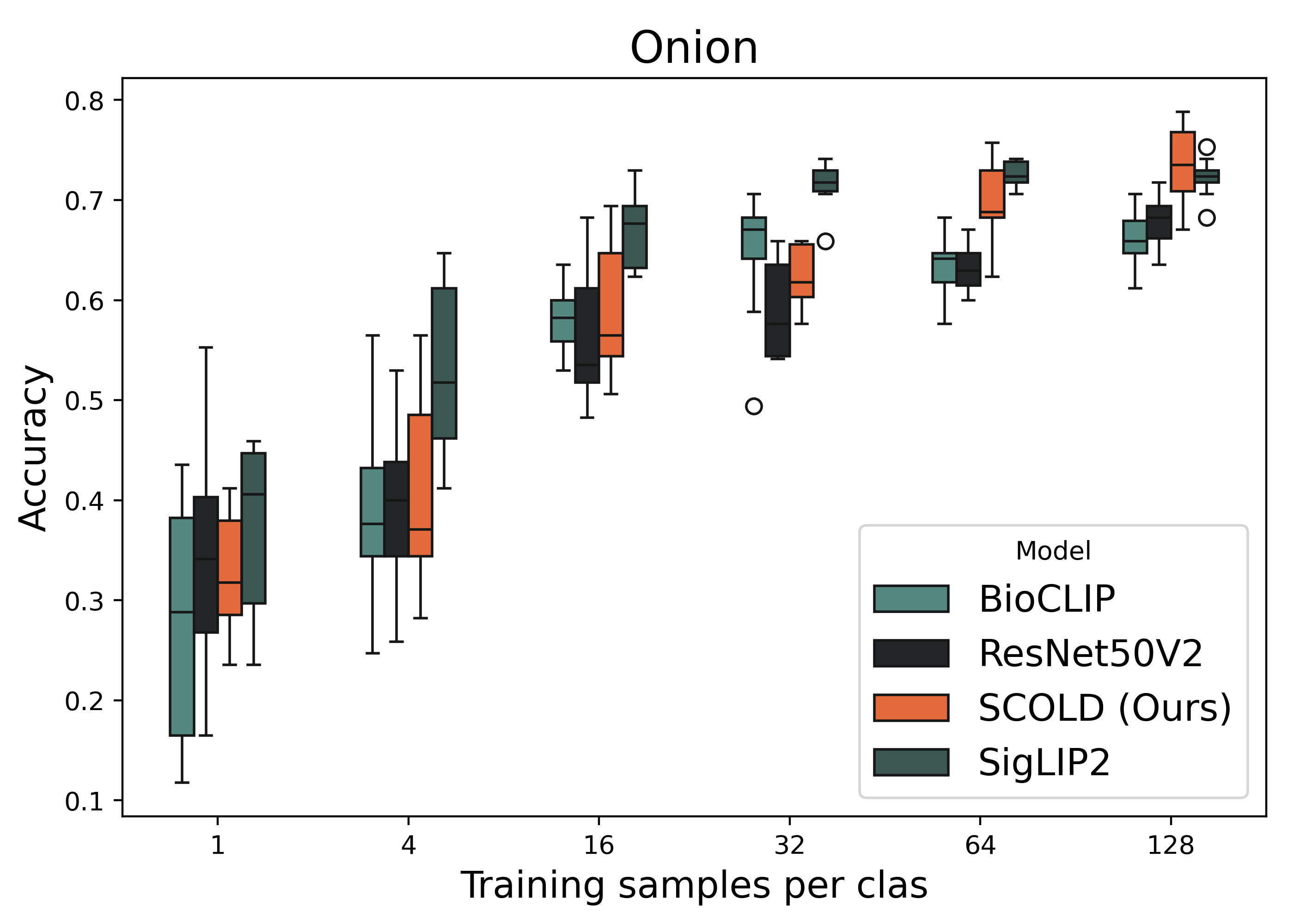}
    \end{subfigure}

    \vspace{0.2cm}

    \begin{subfigure}[b]{0.65\textwidth}
        \centering
        \includegraphics[width=\linewidth]{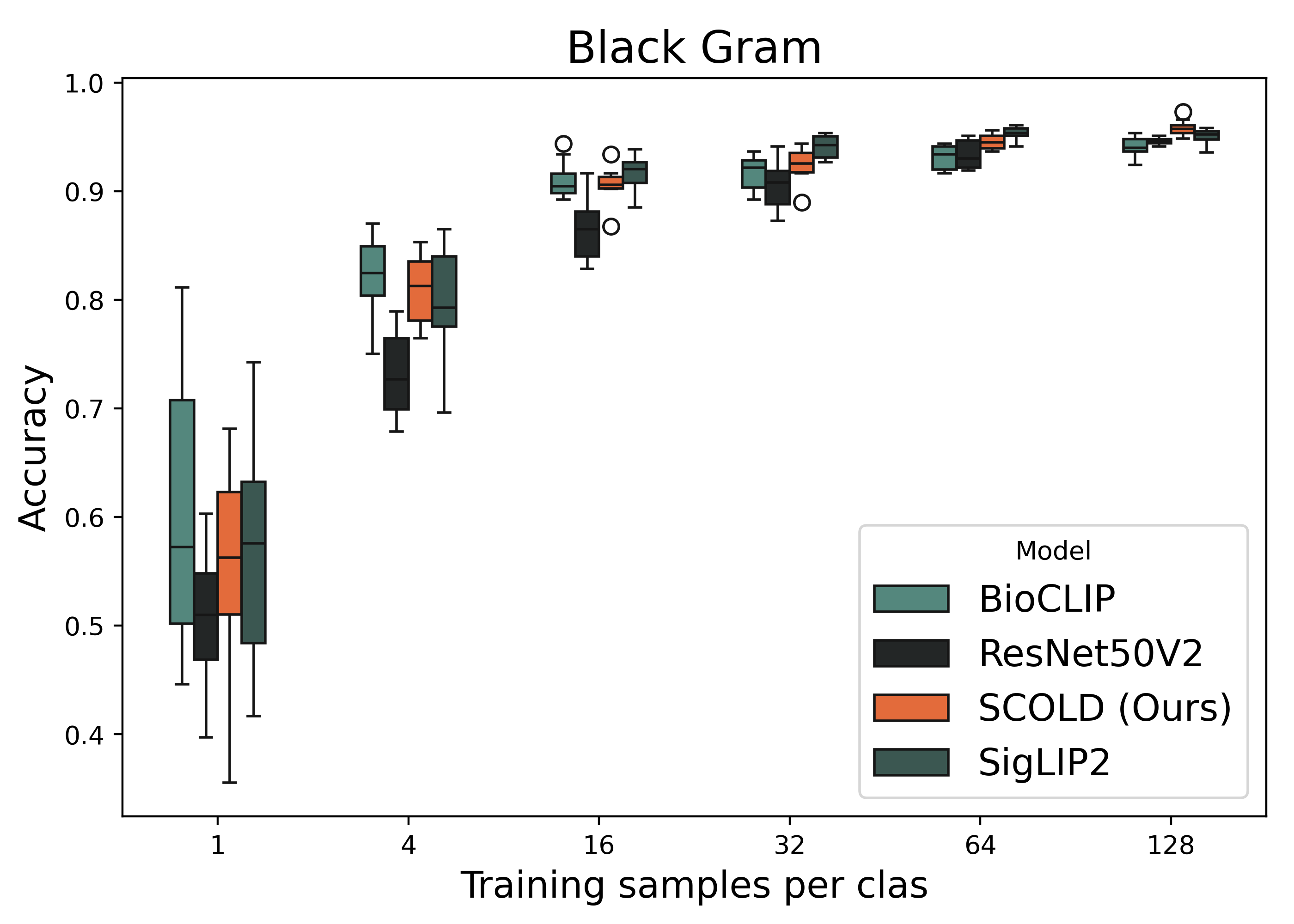}
    \end{subfigure}
    \hfill
    \begin{subfigure}[b]{0.65\textwidth}
        \centering
        \includegraphics[width=\linewidth]{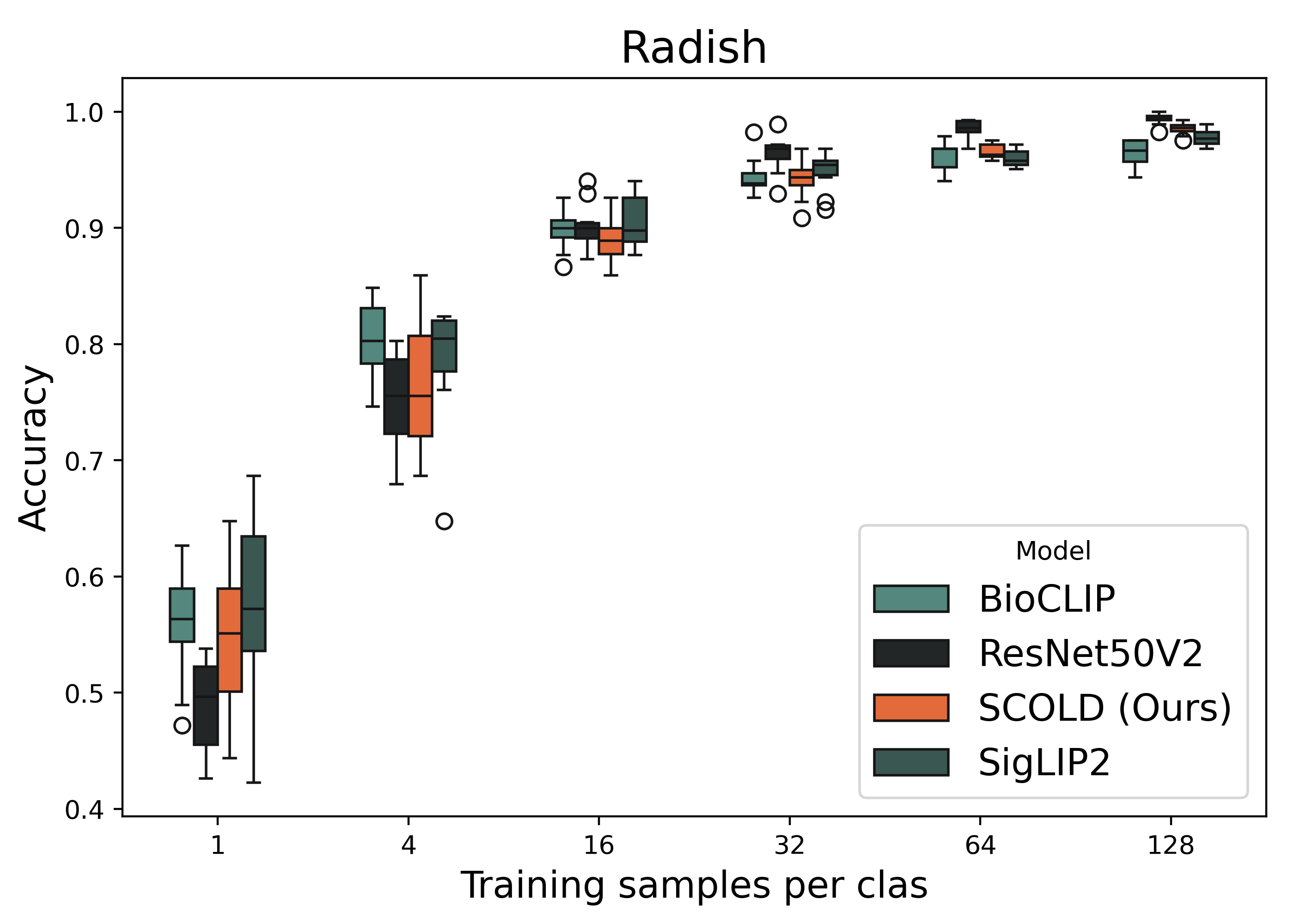}
    \end{subfigure}

    \caption{Comparison of few-shot classification accuracy (\%) via linear probing across various datasets using different models.}
    \label{fig12}
\end{figure}
\end{landscape}

In the linear probing few-shot evaluation, SCOLD exhibited stable and competitive performance across all shot settings, as visualized in the box plots (Figure \ref{fig12}). At the 1-shot level, SCOLD achieved a little lower than other model, due to data limitations of pre-training dataset. As the number of shots increased, the performance of SCOLD consistently improved. Notably, at 16-shot and 32-shot, SCOLD maintained high median accuracies in four dataset, respectively, closely matching or surpassing BioCLIP and SigLIP2. At 128-shot, SCOLD achieved the highest accuracy of in 3 of 4 datasets, reflecting its robust generalization and fine-grained feature representation. Compared to other models, SCOLD also demonstrated a narrower variance at higher shot counts, further emphasizing its reliability and stability in few-shot settings.

\subsection{Image-Text Retrieval }
Next, we validated the efficacy of the proposed method for image-text retrieval tasks. We conducted image text retrieval on the LeafNet testing set for this evaluation. The experimental results are listed in Table \ref{tab5a}, \ref{tab5b}. These demonstrate that \gls{SCOLD} confers vital improvements to LeafNet. The results are summarised regarding the retrieval accuracy at different ranks: R@1, R@5, and R@10. For the Image-to-Text task, \gls{SCOLD} achieved an R@1 value of 95.49\%, outperforming \gls{CLIP} (our implementation) by 0.48\%. The model also showed a slight improvement in R@5 and R@10, with R@5 reaching 98.85\% (+0.96\%) and R@10 reaching 99.93\% (+0.08\%). In the text-to-image task, \gls{SCOLD} delivered performance comparable to \gls{CLIP}, with an R@1 of 95.46\% (+0.15\% over \gls{CLIP}) and a notable improvement in R@5, achieving 99.7\% (+1.52\%). At R@10, \gls{SCOLD} outperformed \gls{CLIP} by a margin of 0.1\%, reaching 99.99\%. These results indicate that \gls{SCOLD} outperforms \gls{CLIP} in the image-to-text tasks while showing competitive performance in text-to-image retrieval.

\begin{table*}[t]
\begin{center}
\caption{\gls{SCOLD} image-to-text retrieval performance on LeafNet.}
\label{tab5a}
\begin{tabular}{|c|ccc|}
\hline
\multirow{2}{*}{\textbf{Method}} & \multicolumn{3}{c|}{\textbf{Image-to-Text}} \\ \cline{2-4} 
                                 & \multicolumn{1}{c|}{R @ 1(\%)}                 & \multicolumn{1}{c|}{R @ 5(\%)}               & R @ 10 (\%)               \\ \hline
\gls{CLIP}                      & \multicolumn{1}{c|}{95.01}                 & \multicolumn{1}{c|}{97.89}               & 99.85                \\ \hline
\gls{SCOLD}                     & \multicolumn{1}{c|}{\textbf{95.49{\color[HTML]{036400}(+0.48)}}} & \multicolumn{1}{c|}{\textbf{98.85{\color[HTML]{036400}(+0.96)}}} & \textbf{99.93{\color[HTML]{036400}(+0.08)}} \\ \hline
\end{tabular}
\end{center}
\end{table*}

\begin{table*}[t]
\begin{center}
\caption{\gls{SCOLD} text-to-image retrieval performance on LeafNet.}
\label{tab5b}
\begin{tabular}{|c|ccc|}
\hline
\multirow{2}{*}{\textbf{Method}} & \multicolumn{3}{c|}{\textbf{Text-to-Image}} \\ \cline{2-4} 
                                 & \multicolumn{1}{c|}{R @ 1(\%)}                 & \multicolumn{1}{c|}{R @ 5(\%)}               & R @ 10(\%)               \\ \hline
\gls{CLIP}                      & \multicolumn{1}{c|}{95.31}                 & \multicolumn{1}{c|}{98.18}               & 99.89                \\ \hline
\gls{SCOLD}                     & \multicolumn{1}{c|}{\textbf{95.46{\color[HTML]{036400}(+0.15)}}} & \multicolumn{1}{c|}{\textbf{99.7{\color[HTML]{036400}(+1.52)}}} & \textbf{99.99{\color[HTML]{036400}(+0.1)}} \\ \hline
\end{tabular}
\end{center}
\end{table*}

\subsection{Image Classification}
\begin{figure*}[!ht]
\centering
\includegraphics[width=1\textwidth]{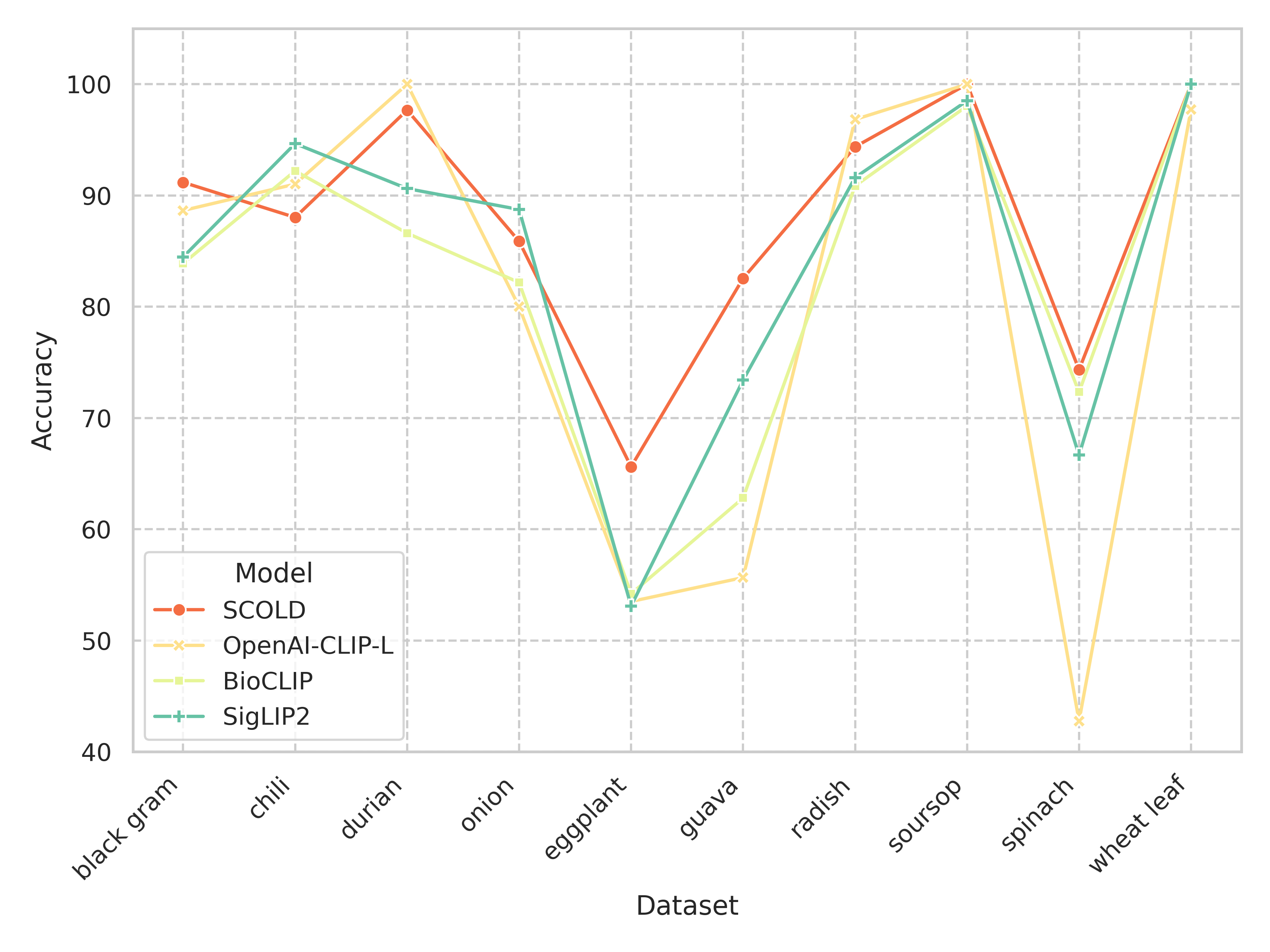}
\caption{Comparison the performance of fine-tuning models on the image classification task.}
\label{fig81}
\end{figure*}

As shown in Figure~\ref{fig81}, \gls{SCOLD} consistently achieves top accuracy across most categories, demonstrating strong generalization after just 8 epochs of fine-tuning. While all models perform competitively on well-represented crops like Durian, Chili, and Radish, \gls{SCOLD} maintains a performance edge. For instance, \gls{SCOLD} scores 97.67\% on Durian and 94.37\% on Radish, closely followed by \gls{CLIP} and SigLIP2.

However, in more challenging datasets such as Eggplant, Guava, and Spinach, \gls{SCOLD} shows clear superiority—achieving 65.62\%, 82.55\%, and 74.34\%, respectively—surpassing CLIP’s 53.51\%, 55.66\%, and 42.76\%. These results suggest that domain-specific pretraining significantly boosts feature discrimination in harder tasks where generalist models struggle. Moreover, \gls{SCOLD} reaches 91.18\% on Black Gram, 85.88\% on Onion, and a perfect 100\% on both Soursop and Wheat Leaf, reinforcing its effectiveness in fine-grained agricultural classification. This highlights the critical role of specialized pre-training for domain adaptation, especially in settings with limited text-image alignment.

\subsection{Ablation Study}
In this section, we first describe the ablation studies conducted to demonstrate the effectiveness of each module in \gls{SCOLD} and then explore several feature representations in the latent space. All ablation experiments were conducted using LeafNet for 12 epochs with a batch size of 16.

\textbf{Effectiveness of $T$ Template and CST:} To verify the effectiveness of long-context prompting and contrastive supervision tuning (CST) for the \gls{SCOLD} model, we conducted ablation studies with three configurations: using only the $T$ template, using only CST, and using both. As shown in Table~\ref{tab8}, incorporating both $T$ templates and CST leads to the highest performance across all recall metrics on the LeafNet dataset, achieving 95.49\% R@1, 99.88\% R@5, and 99.94\% R@10. Using CST alone improves R@1 from 93.72\% to 94.49\%, while using only the $T$ template slightly reduces R@1 to 93.11\% but increases R@5 to 99.83\%. These results demonstrate that CST consistently enhances performance, and its combination with $T$ templates provides complementary benefits. Overall, these findings validate the effectiveness of long-context prompting and contrastive supervision for improving fine-grained recognition in agricultural domains.

\begin{table*}
\begin{center}
\caption{Ablation studies for the loss function with CST and long context on LeafNet dataset (\%).}
\label{tab8}
\begin{tabular}{|l|c|c|c|c|c|}
\hline
\textbf{Method}        & \textbf{$T$ template} & \textbf{CST} & \textbf{R@1} & \textbf{R@5} & \textbf{R@10 } \\ \hline
SCOLD  &                       &              & 93.72              & 98.83                  & 99.05                      \\ \hline
SCOLD        &                       & $\checkmark$            & 94.49              & 99.32                  & 99.88                      \\ \hline
SCOLD                  & $\checkmark$                     &              & 93.11              & 99.83                  & 99.44                      \\ \hline
\rowcolor{cyan!10} \textbf{SCOLD}    &  $\checkmark$                     &  $\checkmark$           & \textbf{95.49}     & \textbf{99.88}         & \textbf{99.94}             \\ \hline
\end{tabular}
\end{center}
\end{table*}

\begin{figure*}[!ht]
\centering
\includegraphics[width=0.8\textwidth]{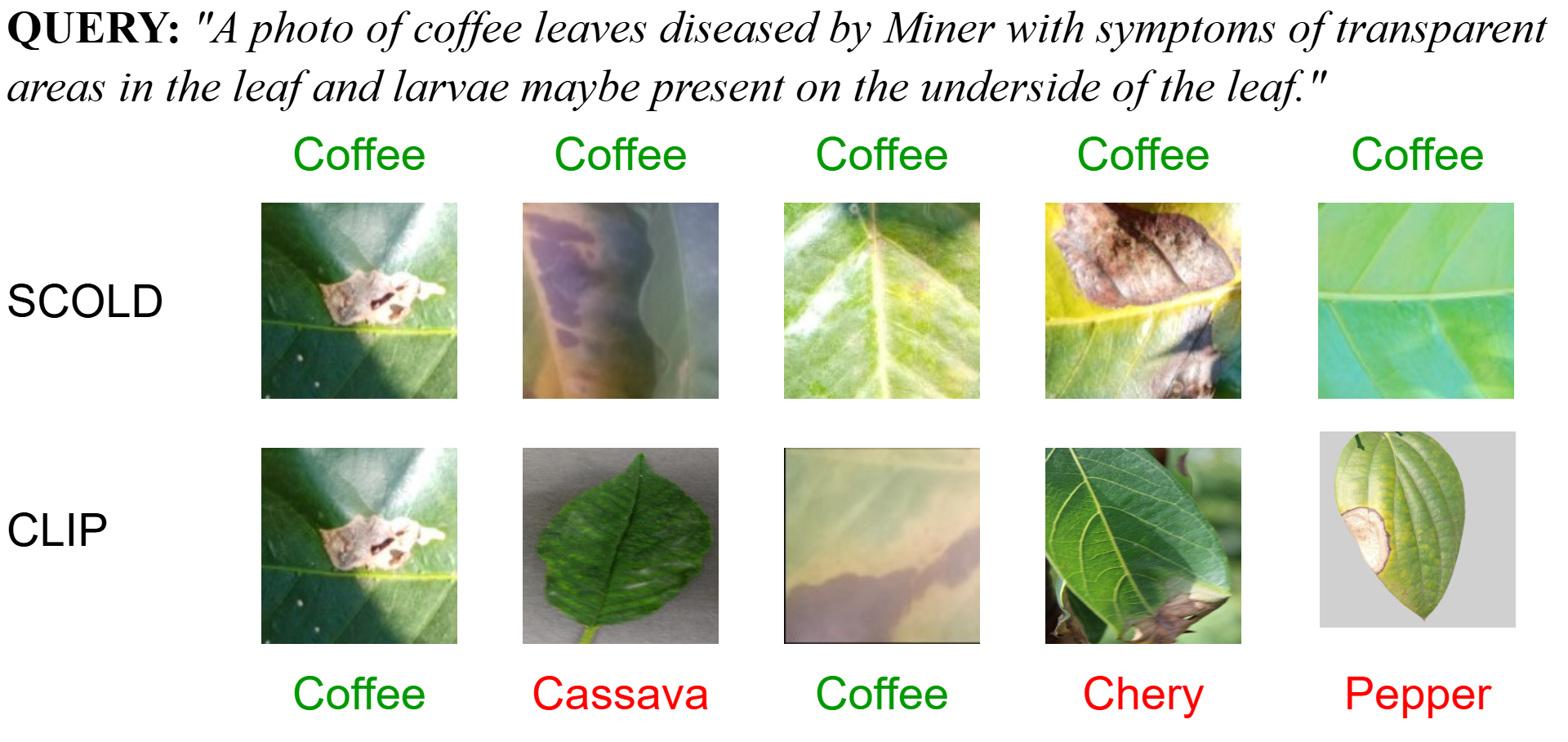}
\caption{Text-to-image Retrieval Top-5 Ranking Result with sample query: “Coffee Leaf Miner” \gls{SCOLD} and \gls{CLIP} models.}
\label{fig9}
\end{figure*}

Moreover, the \gls{CST} technique helps the model perform better and more reliably for image-text retrieval tasks, especially in the case of large-scale datasets. Figure \ref{fig9} demonstrates this reliability when the probability logits obtained by \gls{SCOLD} and \gls{CLIP} without \gls{CST} are entirely different regarding top-5 matching. It can be understood that when we consider one side of the \textit{crop name or disease name} in the query, for example, \textit{apple}, the model with \gls{CST} prioritises the top 10 classes regarding apples first and then the other classes. This also partly helps the model achieve higher efficiency in the downstream task of fine-grained classification and proves that \gls{SCOLD} better understand the text-image relationship and is more effective in matching queries.

\begin{figure*}[!ht]
\centering
\includegraphics[width=1.0\textwidth]{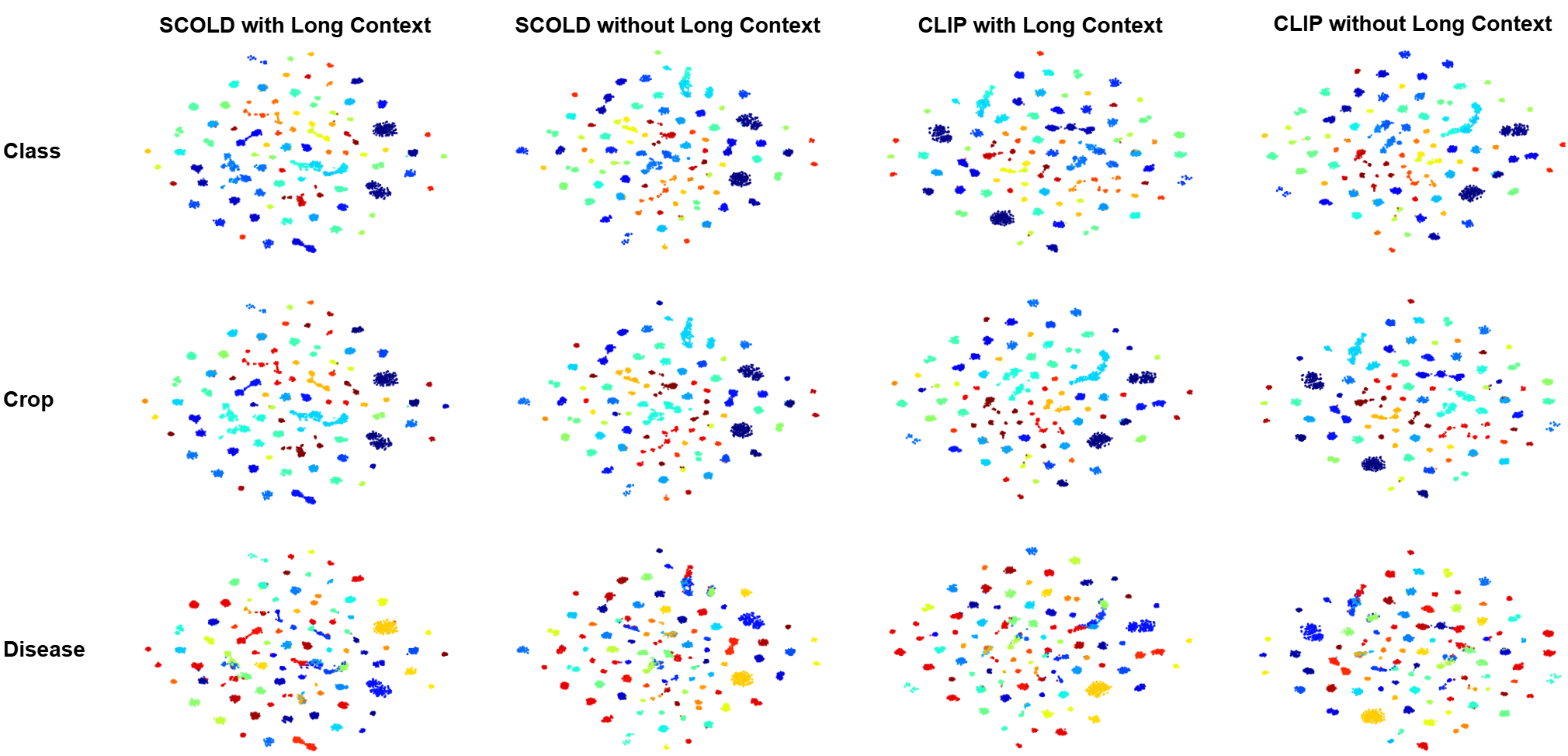}
\caption{t-SNE visualization of ablation experiments models labelling by classes, crops, and diseases.}
\label{fig10}
\end{figure*}

\begin{figure*}[!ht]
\centering
\includegraphics[width=1.0\textwidth]{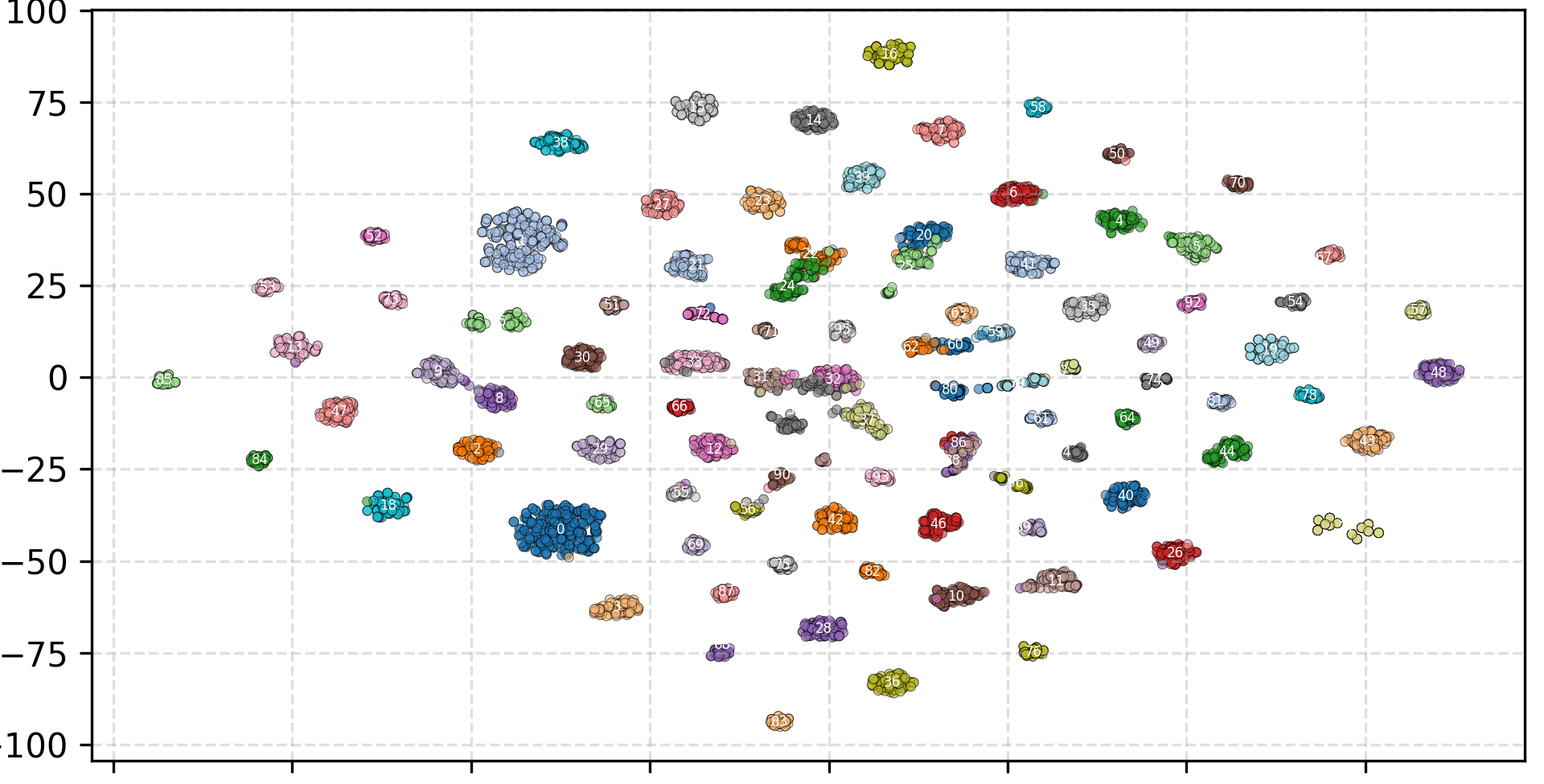}
\caption{t-SNE of features extracted by \gls{SCOLD} model with a long context, labelled by classes.}
\label{fig11}
\end{figure*}

\textbf{Feature Visualization:} The image features were visualised using t-SNE to understand the spatial division between classes according to each tested model. Figure \ref{fig10} also demonstrates that the clustering of each crop type of the \gls{SCOLD} model is the clearest and not as dispersed as in the other models. The \gls{SCOLD} model with a long context in Figure \ref{fig11}. 

At the same time, Table \ref{tab7} also shows that the Silhouette Score\citep{ref46} value of \gls{SCOLD} according to crop and disease and on all classes reaches the highest, especially clustering according to crop, reaching -0.1025 with a long context outperforming all other models. This shows that the information about the correlated features is well utilised by the \gls{SCOLD} model when they are often close to each other, and this helps the model easily achieve a higher performance than the \gls{CLIP} model. 

\begin{table*}
\begin{center}
\caption{Silhouette Score of t-SNE visualization of ablation experiments model by crop, disease, and class.}
\label{tab7}
\begin{tabular}{|cc|p{3cm}|p{3cm}|p{3cm}|}
\hline
\multicolumn{2}{|c|}{\textbf{Method}} & \textbf{\sloppy Cluster by Classes} & \textbf{\sloppy Cluster by Crop} & \textbf{\sloppy Cluster by Disease} \\ \hline
\multicolumn{1}{|c|}{\gls{CLIP}}    & W     & 0.7179                        & -0.185                     & 0.1811                      \\ \hline
\multicolumn{1}{|c|}{\gls{CLIP}}    & W/o   & 0.7257                        & -0.1968                    & 0.1722                      \\ \hline
\multicolumn{1}{|c|}{\gls{SCOLD}}   & W     & \textbf{0.7314}               & \textbf{-0.1025}           & \textbf{0.1955}             \\ \hline
\multicolumn{1}{|c|}{\gls{SCOLD}}   & W/o   & 0.7241                        & -0.1432                    & 0.1745                      \\ \hline
\end{tabular}
\end{center}
\end{table*}

\section{DISCUSSION}
The results from the downstream tasks in Tables \ref{tab3}, \ref{tab4}, \ref{tab5a} and \ref{tab5b}, including zero-shot classification, few-shot classification, and image-text retrieval, highlight \gls{SCOLD}'s superior performance and generalisation ability. Across all the evaluated \gls{OOD} datasets, \gls{SCOLD} with the lower setting model consistently outperformed \gls{CLIP}, SigLIP2 and BioCLIP, with notable accuracy and retrieval precision improvements. These results demonstrate the effectiveness of the proposed framework in leveraging the complementary strengths of visual and textual encoders to improve feature representation and task-specific performance. 

For zero-shot and few-shot classification, \gls{SCOLD} demonstrated robust performance in low-shot scenarios, reflecting its capacity to generalise even with limited labelled data. These findings validate \gls{SCOLD} as a powerful model for real-world applications where labelled data are scarce, particularly for agricultural tasks. Moreover, incorporating long-context templates significantly enhances \gls{SCOLD}'s ability to learn nuanced relationships between images and their corresponding textual descriptions. The model better aligns modalities by extending the textual prompts with the template. This improvement was evident in the zero-shot classification results, where a long context contributed to \gls{SCOLD} outperforming \gls{CLIP} on challenging datasets. 

However, we observed diminishing returns for a long context on certain datasets, such as ours, where the difference between long and simple contexts was minimal. This suggests that while a long context significantly benefits challenging scenarios, its impact may vary depending on the dataset's complexity and semantic requirements.

Another reason for the high performance of \gls{SCOLD} is that the context-aware soft-target mechanism is critical, particularly in tasks involving class similarity or ambiguity. Unlike traditional hard labels that treat all negative pairs equally, our soft labelling approach introduces a weighting scheme based on contextual relevance. This design ensures that semantically similar negatives are treated proportionally, leading to more robust embedding.

\section{Conclusion}\label{Conclusion}
This paper introduced \gls{SCOLD}, a vision-language foundation model tailored for leaf disease identification. We demonstrated its efficacy through extensive experiments across multiple downstream tasks, including zero-shot classification, few-shot classification, and image-text retrieval. Concurrently, we also carefully collected and proposed the LeafNet dataset as the foundation for plant leaf diseases. Moreover, the \gls{SCOLD} model pre-trained on LeafNet achieved more 95\% recall on image text retrieval tasks on the testing set and achieved more 99\% recall in the top 5, with a total of 97 tested classes. This shows the importance of the Soft Target for Contrastive Language-Image Pre-training model with data having a certain level of correlation between classes when \gls{SCOLD} performs better and becomes competitive with the \gls{CLIP}, SigLIP2 and BioCLIP models with the lower parameters and \gls{SOTA} model pre-trained on ImageNet-21K. In downstream tasks, such as zero-shot and few-classification, \gls{SCOLD} also achieved stable performance on \gls{OOD} datasets, especially in 16-shot  mode and fine-tuning, which almost always achieves a performance $>$90\%, notably 98\% on the Durian, Wheat and Soursop leaf disease dataset. Accordingly, this study also proposed detailed descriptions of disease indicators on the leaves of common plants. These descriptions or long contexts also improved the model when recognising in zero-shot mode, when more information from text helps the model focus on features from completely new data and provides more stable diagnosis results. Hence, \gls{SCOLD} and LeafNet are the first vision-language model and data combination capable of foundation-modelling leaf disease classification. In future studies, we plan to expand the research directions, such as object detection and segmentation combined with text features, and collect more multimodal agriculture data to create a foundation model for smart agriculture.

\section*{Acknowledgement}
This work has been submitted to Expert Systems With Applications and is under review.


\begin{thebibliography}{99}


\bibitem[Kok et al.(2021)]{ref01}
 Zhi Hong Kok,  Abdul Rashid Mohamed Shariff,  Meftah Salem M. Alfatni, \&  Siti Khairunniza-Bejo (2021).
Support Vector Machine in Precision Agriculture: A review. 
\textit{Computers and Electronics in Agriculture}, \textit{191}, 106546. \url{http://dx.doi.org/10.1016/j.compag.2021.106546}

\bibitem[Chin et al.(2023)]{ref02}
 Ruben Chin,  Cagatay Catal, \&  Ayalew Kassahun (2023).
Plant disease detection using drones in precision agriculture. 
\textit{Precision Agriculture}, \textit{24}(5), 1663–1682. \url{http://dx.doi.org/10.1007/s11119-023-10014-y}

\bibitem[Coulibaly et al.(2022)]{ref03}
 Solemane Coulibaly,  Bernard Kamsu-Foguem,  Dantouma Kamissoko, \&  Daouda Traore (2022).
Deep learning for precision agriculture: A bibliometric analysis. 
\textit{Intelligent Systems with Applications}, \textit{16}, 200102. \url{http://dx.doi.org/10.1016/j.iswa.2022.200102}

\bibitem[Abd Algani et al.(2023)]{ref04}
 Yousef Methkal Abd Algani,  Orlando Juan Marquez Caro,  Liz Maribel Robladillo Bravo,  Chamandeep Kaur,  Mohammed Saleh Al Ansari, \&  B. Kiran Bala (2023).
Leaf disease identification and classification using optimized deep learning. 
\textit{Measurement: Sensors}, \textit{25}, 100643. \url{http://dx.doi.org/10.1016/j.measen.2022.100643}

\bibitem[Quach et al.(2023)]{ref05}
Quach, L.-D., Nguyen, Q. A., Nguyen, Q. K., \& Nguyen, T. N. (2023). 
Using optimization algorithm to improve the accuracy of the CNN model on the rice leaf disease dataset. 
\textit{Information Systems for Intelligent Systems} (pp. 535–544). Springer Nature Singapore. \url{https://doi.org/10.1007/978-981-19-7447-2_47}

\bibitem[Zeng et al.(2022)]{ref06}
 Weihui Zeng,  Haidong Li,  Gensheng Hu, \&  Dong Liang (2022).
Lightweight dense-scale network (LDSNet) for corn leaf disease identification. 
\textit{Computers and Electronics in Agriculture}, \textit{197}, 106943. \url{http://dx.doi.org/10.1016/j.compag.2022.106943}

\bibitem[Khan et al.(2022)]{ref07}
 Asif Iqbal Khan,  S.M.K. Quadri,  Saba Banday, \&  Junaid Latief Shah (2022).
Deep diagnosis: A real-time apple leaf disease detection system based on deep learning. 
\textit{Computers and Electronics in Agriculture}, \textit{198}, 107093. \url{http://dx.doi.org/10.1016/j.compag.2022.107093}

\bibitem[Paymode et al.(2022)]{ref08}
 Ananda S. Paymode \&  Vandana B. Malode (2022).
Transfer Learning for Multi-Crop Leaf Disease Image Classification using Convolutional Neural Network VGG. 
\textit{Artificial Intelligence in Agriculture}, \textit{6}, 23–33. \url{http://dx.doi.org/10.1016/j.aiia.2021.12.002}

\bibitem[Vo et al.(2023)]{ref09}
 Hoang-Tu Vo,  Luyl-Da Quach, \&  Hoang Tran Ngoc (2023).
Ensemble of Deep Learning Models for Multi-plant Disease Classification in Smart Farming. 
\textit{International Journal of Advanced Computer Science and Applications}, \textit{14}(5), 155–163. \url{http://dx.doi.org/10.14569/IJACSA.2023.01405108}

\bibitem[Nguyen et al.(2024)]{ref10}
Nguyen, A. Q., Tran, M. T., Nguyen, Q. N., Huynh, H. K., Le, L. T. T., \& Quach, L.-D. (2024).
Classification of rice plant disease based on descriptive information with DistilBERT’s architecture.
In *Proceedings of the 2024 9th International Conference on Intelligent Information Technology* (pp. 155–163). ACM. \url{https://doi.org/10.1145/3654522.3654568}

\bibitem[Quach et al.(2023)]{ref11}
 Luyl-Da Quach,  Anh Nguyen Quynh,  Khang Nguyen Quoc, \&  An Nguyen Thi Thu (2023).
Using the Term Frequency-Inverse Document Frequency for the Problem of Identifying Shrimp Diseases with State Description Text. 
\textit{International Journal of Advanced Computer Science and Applications}, \textit{14}(5), . \url{http://dx.doi.org/10.14569/IJACSA.2023.0140577}

\bibitem[Zhang et al.(2024)]{ref12}
 Kunpeng Zhang,  Li Ma,  Beibei Cui,  Xin Li,  Boqiang Zhang, \&  Na Xie (2024).
Visual large language model for wheat disease diagnosis in the wild. 
\textit{Computers and Electronics in Agriculture}, \textit{227}, 109587. \url{http://dx.doi.org/10.1016/j.compag.2024.109587}

\bibitem[Zhu et al.(2025)]{ref13}
 Hongfei Zhu,  Weiming Shi,  Xinyu Guo,  Shiting Lyu,  Ranbing Yang, \&  Zhongzhi Han (2025).
Potato disease detection and prevention using multimodal AI and large language model. 
\textit{Computers and Electronics in Agriculture}, \textit{229}, 109824. \url{http://dx.doi.org/10.1016/j.compag.2024.109824}

\bibitem[Liaw et al.(2024)]{ref14}
Liaw, J. Z., Chai, A. Y. H., Lee, S. H., Bonnet, P., \& Joly, A. (2024).
Can language improve visual features for distinguishing unseen plant diseases?
In *Pattern Recognition* (pp. 296–311). Springer Nature Switzerland. \url{https://doi.org/10.1007/978-3-031-78113-1_20}

\bibitem[Sarkar et al.(2023)]{ref15}
 Chittabarni Sarkar,  Deepak Gupta,  Umesh Gupta, \&  Barenya Bikash Hazarika (2023).
Leaf disease detection using machine learning and deep learning: Review and challenges. 
\textit{Applied Soft Computing}, \textit{145}, 110534. \url{https://doi.org/10.48550/ARXIV.1807.03748}

\bibitem[van den Oord et al.(2018)]{ref16}
van den Oord, A., Li, Y., \& Vinyals, O. (2018).
Representation learning with contrastive predictive coding.
arXiv. \url{https://doi.org/10.48550/ARXIV.1807.03748}

\bibitem[Wei et al.(2022)]{ref17}
Wei, Y., Cao, Y., Zhang, Z., Yao, Z., Xie, Z., Hu, H., \& Guo, B. (2022).
iCAR: Bridging image classification and image-text alignment for visual recognition.
arXiv. \url{https://doi.org/10.48550/ARXIV.2204.10760}

\bibitem[Yang et al.(2022)]{ref18}
Yang, J., Li, C., Zhang, P., Xiao, B., Liu, C., Yuan, L., \& Gao, J. (2022).
Unified contrastive learning in image-text-label space.
In *2022 IEEE/CVF Conference on Computer Vision and Pattern Recognition (CVPR)* (pp. 19141–19151). IEEE. \url{https://doi.org/10.1109/CVPR52688.2022.01857}

\bibitem[Wei et al.(2023)]{ref19}
Wei, Y., Cao, Y., Zhang, Z., Peng, H., Yao, Z., Xie, Z., Hu, H., \& Guo, B. (2023).
iCLIP: Bridging image classification and contrastive language-image pre-training for visual recognition.
In *2023 IEEE/CVF Conference on Computer Vision and Pattern Recognition (CVPR)* (pp. 2776–2786). IEEE. \url{https://doi.org/10.1109/CVPR52729.2023.00272}


\bibitem[Zhou et al.(2024)]{ref20}
 Yueyue Zhou,  Hongping Yan,  Kun Ding,  Tingting Cai, \&  Yan Zhang (2024).
Few-Shot Image Classification of Crop Diseases Based on Vision–Language Models. 
\textit{Sensors}, \textit{24}(18), 6109. \url{http://dx.doi.org/10.3390/s24186109}

\bibitem[Quach et al.(2023)]{ref21}
 Luyl-Da Quach,  Khang Nguyen Quoc,  Anh Nguyen Quynh,  Nguyen Thai-Nghe, \&  Tri Gia Nguyen (2023).
Explainable Deep Learning Models With Gradient-Weighted Class Activation Mapping for Smart Agriculture. 
\textit{IEEE Access}, \textit{11}, 83752–83762. \url{http://dx.doi.org/10.1109/ACCESS.2023.3296792}

\bibitem[Quach et al.(2024)]{ref22}
Quach, L.-D., Nguyen, Q. K., Nguyen, C.-N., \& Nguyen, T.-N. (2024).
Explainable AI for plant disease detection: Assessing explainability in classifying maize leaves diseases with Focus Score and Ablation-CAM.
In *Intelligent Systems and Data Science* (pp. 19–32). Springer Nature Singapore. \url{https://doi.org/10.1007/978-981-97-9613-7_2}

\bibitem[Quach et al.(2025)]{ref23}
 Luyl-Da Quach,  Nguyen Quoc Khang,  Nguyen Thai-Nghe, \&  Chi-Ngon Nguyen (2025).
XAI-BO: an architecture using Grad-CAM technique to evaluate Bayesian optimization algorithms on deep learning models. 
\textit{Journal of Information and Telecommunication}, \textit{}, 1–22. \url{https://doi.org/10.48550/ARXIV.2103.00020}

\bibitem[Radford et al.(2021)]{ref24}
Radford, A., Kim, J. W., Hallacy, C., Ramesh, A., Goh, G., Agarwal, S., ... \& Sutskever, I. (2021).
Learning transferable visual models from natural language supervision.
arXiv. \url{https://doi.org/10.48550/ARXIV.2103.00020}


\bibitem[Müller et al.(2019)]{ref26}
Müller, R., Kornblith, S., \& Hinton, G. (2019).
When does label smoothing help?
arXiv. \url{https://doi.org/10.48550/ARXIV.1906.02629}

\bibitem[Gao et al.(2024)]{ref27}
 Yuting Gao,  Jinfeng Liu,  Zihan Xu,  Tong Wu,  Enwei Zhang,  Ke Li,  Jie Yang,  Wei Liu, \&  Xing Sun (2024).
SoftCLIP: Softer Cross-Modal Alignment Makes CLIP Stronger. 
\textit{Proceedings of the AAAI Conference on Artificial Intelligence}, \textit{38}(3), 1860–1868. \url{http://dx.doi.org/10.1609/aaai.v38i3.27955}

\bibitem[Jia et al.(2021)]{ref28}
 Chao Jia,  Yinfei Yang,  Ye Xia,  Yi-Ting Chen,  Zarana Parekh,  Hieu Pham,  Quoc V. Le,  Yunhsuan Sung,  Zhen Li, \&  Tom Duerig (2021).
Scaling Up Visual and Vision-Language Representation Learning With Noisy Text Supervision. 
\textit{}, \textit{}, . \url{https://doi.org/10.48550/ARXIV.2102.05918}

\bibitem[G. et al.(2019)]{ref29}
 Geetharamani G. \&  Arun Pandian J. (2019).
Identification of plant leaf diseases using a nine-layer deep convolutional neural network. 
\textit{Computers \& Electrical Engineering}, \textit{76}, 323–338. \url{https://doi.org/10.17632/378C69NWK5.1}

\bibitem[Anan et al.(2024)]{ref30}
Anan, I. J., Ahmmed, T., \& Mojumdar, M. U. (2024).
Grape leaf dataset for leaf health classification and disease detection.
Mendeley Data. \url{https://doi.org/10.17632/378C69NWK5.1}

\bibitem[Thite et al.(2024)]{ref31}
 Sandip Thite,  Yogesh Suryawanshi,  Kailas Patil, \&  Prawit Chumchu (2024).
Sugarcane leaf dataset: A dataset for disease detection and classification for machine learning applications. 
\textit{Data in Brief}, \textit{53}, 110268. \url{http://dx.doi.org/10.1016/j.dib.2024.110268}

\bibitem[Ahmed et al.(2023)]{ref32}
 Sarder Iftekhar Ahmed,  Muhammad Ibrahim,  Md. Nadim,  Md. Mizanur Rahman,  Maria Mehjabin Shejunti,  Taskeed Jabid, \&  Md. Sawkat Ali (2023).
MangoLeafBD: A comprehensive image dataset to classify diseased and healthy mango leaves. 
\textit{Data in Brief}, \textit{47}, 108941. \url{http://dx.doi.org/10.1016/j.dib.2023.108941}

\bibitem[Jepkoech et al.(2021)]{ref33}
 Jennifer Jepkoech,  David Muchangi Mugo,  Benson K. Kenduiywo, \&  Edna Chebet Too (2021).
Arabica coffee leaf images dataset for coffee leaf disease detection and classification. 
\textit{Data in Brief}, \textit{36}, 107142. \url{https://doi.org/10.17632/TT2SMZRZRS.4}

\bibitem[Ahmad(2024)]{ref34}
Ahmad, M. H. (2024).
Advanced tea crop disease study: High-resolution dataset for precision agriculture and pathological insight.
Mendeley Data. \url{https://doi.org/10.17632/TT2SMZRZRS.4}

\bibitem[Kwabena Adu(2023)]{ref35}
Kwabena Adu. (2023).
Dataset for crop pest and disease detection.
Mendeley. \url{https://doi.org/10.17632/BWH3ZBPKPV.1}

\bibitem[Mduma et al.(2023)]{ref36}
 Neema Mduma \&  Hudson Laizer (2023).
Machine Learning Imagery Dataset for Maize Crop: A Case of Tanzania. 
\textit{Data in Brief}, \textit{48}, 109108. \url{http://dx.doi.org/10.1016/j.dib.2023.109108}

\bibitem[Sultana et al.(2023)]{ref37}
 Nusrat Sultana,  Sumaita Binte Shorif,  Morium Akter, \&  Mohammad Shorif Uddin (2023).
A dataset for successful recognition of cucumber diseases. 
\textit{Data in Brief}, \textit{49}, 109320. \url{http://dx.doi.org/10.1016/j.dib.2023.109320}

\bibitem[Quach et al.(2022)]{ref38}
 Luyl-Da Quach,  Khang Nguyen Quoc,  Anh Nguyen Quynh, \&  Hoang Tran Ngoc (2022).
Evaluation of the Efficiency of the Optimization Algorithms for Transfer Learning on the Rice Leaf Disease Dataset. 
\textit{International Journal of Advanced Computer Science and Applications}, \textit{13}(10), 249–253. \url{https://doi.org/10.48550/ARXIV.1907.11692}

\bibitem[Singh et al.(2020)]{ref39}
Singh, D., Jain, N., Jain, P., Kayal, P., Kumawat, S., \& Batra, N. (2020).
PlantDoc: A dataset for visual plant disease detection.
In *Proceedings of the 7th ACM IKDD CoDS and 25th COMAD* (pp. 249–253). ACM. \url{https://doi.org/10.1145/3371158.3371196}

\bibitem[Liu et al.(2021)]{ref40}
Liu, Z., Lin, Y., Cao, Y., Hu, H., Wei, Y., Zhang, Z., Lin, S., \& Guo, B. (2021).
Swin Transformer: Hierarchical vision transformer using shifted windows.
In 2021 IEEE/CVF International Conference on Computer Vision (ICCV)* (pp. 9992–10002). IEEE. \url{https://doi.org/10.1109/ICCV48922.2021.00986}

\bibitem[Liu et al.(2019)]{ref41}
Liu, Y., Ott, M., Goyal, N., Du, J., Joshi, M., Chen, D., ... \& Stoyanov, V. (2019).
RoBERTa: A robustly optimized BERT pretraining approach.
arXiv. \url{https://doi.org/10.48550/ARXIV.1907.11692}

\bibitem[Loshchilov \& Hutter(2017)]{ref42}
Loshchilov, I., \& Hutter, F. (2017).
Decoupled weight decay regularization.
arXiv. \url{https://doi.org/10.48550/ARXIV.1711.05101}

\bibitem[Getachew(2021)]{ref43}
Getachew, H. (2021).
Wheat leaf dataset.
Mendeley. \url{https://doi.org/10.17632/WGD66F8N6H.1}

\bibitem[Aishwarya et al.(2024)]{ref44}
 M. P Aishwarya \&  A. Padmanabha Reddy (2024).
Dataset of chilli and onion plant leaf images for classification and detection. 
\textit{Data in Brief}, \textit{54}, 110524. \url{https://doi.org/10.17632/HJRHRT5HS8.1}

\bibitem[Mustofa et al.(2024)]{ref45}
Mustofa, S., Shakib, M. M. H., \& Emon, Y. R. (2024).
SoursopBD: A medicinal plant leaf disease dataset for machine learning and computer vision tasks.
Mendeley Data. \url{https://doi.org/10.17632/HJRHRT5HS8.1}

\bibitem[Rousseeuw Rousseeuw(1987)]{ref46}
 Peter J. Rousseeuw (1987).
Silhouettes: A graphical aid to the interpretation and validation of cluster analysis. 
\textit{Journal of Computational and Applied Mathematics}, \textit{20}, 53–65. \url{http://dx.doi.org/10.1016/0377-0427(87)90125-7}

\bibitem[Hasan et al.(2025)]{ref47}
 Mahamudul Hasan,  Raiyan Gani,  Mohammad Rifat Ahmmad Rashid,  Maherun Nessa Isty,  Raka Kamara, \&  Taslima Khan Tarin (2025).
Smartphone image dataset for radish plant leaf disease classification from Bangladesh. 
\textit{Data in Brief}, \textit{58}, 111263. \url{http://dx.doi.org/10.1016/j.dib.2024.111263}

\bibitem[Shoib et al.(2025)]{ref48}
 Md. Mehedi Hasan Shoib,  Shahnewaz Saeem,  Afia Benta Aziz Tonima, \&  Mayen Uddin Mojumdar (2025).
IDBGL: A unique image dataset of black gram (Vigna mungo) leaves for disease detection and classification. 
\textit{Data in Brief}, \textit{59}, 111347. \url{http://dx.doi.org/10.1016/j.dib.2025.111347}

\bibitem[Howlader et al.(2025)]{ref49}
 Shakib Howlader,  Md. Sabbir Ahamed,  Mayen Uddin Mojumdar,  Sheak Rashed Haider Noori,  Shah Md Tanvir Siddiquee, \&  Narayan Ranjan Chakraborty (2025).
A comprehensive image dataset for the identification of eggplant leaf diseases and computer vision applications. 
\textit{Data in Brief}, \textit{59}, 111353. \url{http://dx.doi.org/10.1016/j.dib.2025.111353}

\bibitem[Shihab et al.(2025)]{ref50}
 Montasir Rahman Shihab,  Nafiu Islam Saim,  Mayen Uddin Mojumdar,  Dewan Mamun Raza,  Shah Md Tanvir Siddiquee,  Sheak Rashed Haider Noori, \&  Narayan Ranjan Chakraborty (2025).
Image Dataset for Classification of Diseases in Guava Fruits and Leaves. 
\textit{Data in Brief}, \textit{}, 111378. \url{http://dx.doi.org/10.1016/j.dib.2025.111378}

\bibitem[Sayeem et al.(2025)]{ref51}
 Adnan Rahman Sayeem,  Jannatul Ferdous Omi,  Mehedi Hasan,  Mayen Uddin Mojumdar, \&  Narayan Ranjan Chakraborty (2025).
IDDMSLD: An image dataset for detecting Malabar spinach leaf diseases. 
\textit{Data in Brief}, \textit{58}, 111293. \url{http://dx.doi.org/10.1016/j.dib.2025.111293}

\bibitem[Singh et al.(2022)]{ref52}
Singh, A., Hu, R., Goswami, V., Couairon, G., Galuba, W., Rohrbach, M., \& Kiela, D. (2022).
FLAVA: A foundational language and vision alignment model.
In *2022 IEEE/CVF Conference on Computer Vision and Pattern Recognition (CVPR)* (pp. 15617–15629). IEEE. \url{https://doi.org/10.1109/CVPR52688.2022.01519}

\bibitem[Li et al.(2022)]{ref53}
Li, J., Li, D., Xiong, C., \& Hoi, S. (2022).
BLIP: Bootstrapping language-image pre-training for unified vision-language understanding and generation.
In *Proceedings of the 39th International Conference on Machine Learning* (pp. 12888–12900). PMLR.

\bibitem[Sajitha et al.(2024)]{ref54}
 P. Sajitha,  A. Diana Andrushia,  N. Anand, \&  M.Z. Naser (2024).
A review on machine learning and deep learning image-based plant disease classification for industrial farming systems. 
\textit{Journal of Industrial Information Integration}, \textit{38}, 100572. \url{https://doi.org/10.48550/ARXIV.2109.13228}

\bibitem[Asano et al.(2021)]{ref55}
Asano, Y. M., Rupprecht, C., Zisserman, A., \& Vedaldi, A. (2021).
PASS: An ImageNet replacement for self-supervised pretraining without humans.
arXiv. \url{https://doi.org/10.48550/ARXIV.2109.13228}

\bibitem[Taesiri et al.(2023)]{ref56}
Taesiri, M. R., Nguyen, G., Habchi, S., Bezemer, C.-P., \& Nguyen, A. (2023).
ImageNet-Hard: The hardest images remaining from a study of the power of zoom and spatial biases in image classification.
In *NeurIPS 2023* (Art. 1558, pp. 1–76). Curran Associates Inc.

\bibitem[Kornblith et al.(2019)]{ref57}
Kornblith, S., Shlens, J., \& Le, Q. V. (2019).
Do better ImageNet models transfer better?
In *Proceedings of the IEEE/CVF Conference on Computer Vision and Pattern Recognition (CVPR)*.

\bibitem[Tingting Zhang et al.(2025)]{ref58}
Tingting Zhang, Jing Li, Jinpeng Tong, Yihu Song, Li Wang, Renye Wu, Xuan Wei, Yuanyuan Song, \& Rensen Zeng (2025).
End-to-end deep fusion of hyperspectral imaging and computer vision techniques for rapid detection of wheat seed quality. 
\textit{Artificial Intelligence in Agriculture}, \textit{}, . \url{https://doi.org/10.1016/j.aiia.2025.02.003}

\bibitem[Juan Felipe Restrepo-Arias et al.(2024)]{ref59}
Juan Felipe Restrepo-Arias, John W. Branch-Bedoya, \& Gabriel Awad (2024).
Image classification on smart agriculture platforms: Systematic literature review. 
\textit{Artificial Intelligence in Agriculture}, \textit{13}, 1-17. \url{https://doi.org/10.1016/j.aiia.2024.06.002}

\bibitem[Gan Yang et al.(2025)]{ref60}
Gan Yang, Qifeng Li, Chunjiang Zhao, Chaoyuan Wang, Hua Yan, Rui Meng, Yu Liu, \& Ligen Yu (2025).
TGFN-SD: A text-guided multimodal fusion network for swine disease diagnosis. 
\textit{Artificial Intelligence in Agriculture}, \textit{}, . \url{https://doi.org/10.1016/j.aiia.2025.03.002}

\bibitem[Berka et al.(2023)]{ref61}
 Anas Berka,  Adel Hafiane,  Youssef Es-Saady,  Mohamed El Hajji,  Raphaël Canals, \&  Rachid Bouharroud (2023).
CactiViT: Image-based smartphone application and transformer network for diagnosis of cactus cochineal. 
\textit{Artificial Intelligence in Agriculture}, \textit{9}, 12–21. \url{http://dx.doi.org/10.1016/j.aiia.2023.07.002}

\bibitem[Ghazal et al.(2024)]{ref62}
 Sumaira Ghazal,  Arslan Munir, \&  Waqar S. Qureshi (2024).
Computer vision in smart agriculture and precision farming: Techniques and applications. 
\textit{Artificial Intelligence in Agriculture}, \textit{13}, 64–83. \url{http://dx.doi.org/10.1016/j.aiia.2024.06.004}

\bibitem[He et al.(2015)]{ref63}
He, K., Zhang, X., Ren, S., \& Sun, J. (2015). \textit{Deep residual learning for image recognition}. arXiv preprint arXiv:1512.03385. \url{https://doi.org/10.48550/arXiv.1512.03385}

\bibitem[Stevens et al.(2024)]{ref64}
Stevens, S., Wu, J., Thompson, M. J., Campolongo, E. G., Song, C. H., Carlyn, D. E., Dong, L., Dahdul, W. M., Stewart, C., Berger-Wolf, T., Chao, W.-L., \& Su, Y. (2024). \textit{BioCLIP: A vision foundation model for the tree of life}. In \textit{Proceedings of the 2024 IEEE/CVF Conference on Computer Vision and Pattern Recognition (CVPR)} (pp. 19412–19424). IEEE. \url{https://doi.org/10.1109/CVPR52733.2024.01836}

\bibitem[Tschannen et al.(2025)]{ref65}
Tschannen, M., Gritsenko, A., Wang, X., Naeem, M. F., Alabdulmohsin, I., Parthasarathy, N., Evans, T., Beyer, L., Xia, Y., Mustafa, B., Hénaff, O., Harmsen, J., Steiner, A., \& Zhai, X. (2025). \textit{SigLIP 2: Multilingual vision-language encoders with improved semantic understanding, localization, and dense features}. arXiv preprint arXiv:2502.14786. \url{https://doi.org/10.48550/arXiv.2502.14786}

\end{thebibliography}
\end{document}